\documentclass[12pt]{article}% Math and Physical Sciences Numbered Reference Style
\usepackage{etoolbox}
\usepackage{graphicx}%
\usepackage{csquotes}
\usepackage{multirow}%
\usepackage{amsmath,amssymb,amsfonts}%
\usepackage{amsthm}%
\usepackage{mathrsfs}%

\usepackage{subcaption}
\usepackage[title]{appendix}%
\usepackage{textcomp}%
\usepackage{manyfoot}%
\usepackage{booktabs}%
\usepackage{algorithm}%
\usepackage{algorithmicx}%
\usepackage{algpseudocode}%
\usepackage{listings}%
\usepackage{cite}
\usepackage{hyperref}
\usepackage{booktabs, adjustbox}
\usepackage{float}
\usepackage{bm}
\usepackage{caption}
\usepackage{soul}

\usepackage{amsmath,amssymb,amsfonts}
\usepackage{graphicx}

\newtheorem{theorem}{Theorem}%  meant for continuous numbers
\newtheorem{definition}{Definition}%

\title{Jordan-Segmentable Masks: A Topology-Aware definition for characterizing Binary Image Segmentation}

\author{
Serena Grazia De Benedictis*, Amedeo Altavilla*,\\ Nicoletta Del Buono* \\
\small *Department of Mathematics, University of Bari Aldo Moro, \\ \small Edoardo Orabona street 4, 70125 Bari, Italy\\
\small \texttt{\{serena.debenedictis, amedeo.altavilla, nicoletta.delbuono\}@uniba.it}
}

\date{}

\begin{document}



\maketitle

\begin{abstract}
Image segmentation plays a central role in computer vision. However, widely used evaluation metrics, whether pixel-wise, region-based, or boundary-focused, often struggle to capture the structural and topological coherence of a segmentation. In many practical scenarios, such as medical imaging or object delineation, small inaccuracies in boundary, holes, or fragmented predictions can result in high metric scores, despite the fact that the resulting masks fail to preserve the object global shape or connectivity. This highlights a limitation of conventional metrics: they are unable to assess whether a predicted segmentation partitions the image into meaningful interior and exterior regions.

In this work, we introduce a topology-aware notion of segmentation based on the Jordan Curve Theorem, and adapted for use in digital planes. We define the concept of a \emph{Jordan-segmentatable mask}, which is a binary segmentation whose structure ensures a topological separation of the image domain into two connected components. We analyze segmentation masks through the lens of digital topology and homology theory, extracting a $4$-curve candidate from the mask, verifying its topological validity using Betti numbers. A mask is considered Jordan-segmentatable when this candidate forms a digital 4-curve with $\beta_0 = \beta_1 = 1$, or equivalently when its complement splits into exactly two $8$-connected components.

This framework provides a mathematically rigorous, unsupervised criterion with which to assess the structural coherence of segmentation masks. By combining digital Jordan theory and homological invariants, our approach provides a valuable alternative to standard evaluation metrics, especially in applications where topological correctness must be preserved.
\end{abstract}




\section{Introduction}\label{introduction}

Image segmentation is a fundamental task in computer vision and image processing. It involves partitioning an image into meaningful, homogeneous regions. The objective is to assign each pixel to a particular region or category, which usually represents objects, boundaries, or other significant structures within the image. Specifically, binary segmentation divides the image into two disjoint regions: the foreground (ideally a section to be highlighted) and the background. This process is important in applications where it is critical to isolate a target object from irrelevant regions, such as object detection, medical image analysis and scene understanding \cite{binary_segmentation,introduz1,introduz2,introduz3,introduz4,introduz5,introduz6,introduz7}.

Image segmentation method evaluation can generally be divided into two main paradigms: supervised and unsupervised approaches.  
Supervised evaluation uses reference images (ground truth) to quantify the agreement between predicted and true segmentations. In contrast, unsupervised evaluation assesses intrinsic properties of the segmented regions, such as homogeneity, contrast, or statistical uniformity, without requiring ground truth \cite{metriche,metriche2}. Various metrics have been proposed within these paradigms, including pixel-level measures, region-level consistency measures, and boundary alignment measures. Despite this variety, the most common adopted metrics in practice are the pixel-wise ones, due to their simplicity, interpretability, and compatibility with pixel-annotated datasets. This makes them particularly suitable for benchmarking and method comparison. 

All conventional segmentation metrics — pixel-wise, region-based, and boundary-based have some limitations in capturing the spatial structure and topological characteristics of segmented regions. Consequently, they may not accurately reflect the quality of segmentation in scenarios where topological information plays a significant role. They often overlook aspects such as geometric coherence or topological connectivity in the predicted masks. Minor boundary shifts, small holes, or fragmented regions can sometimes result in relatively high scores, even if the segmentation does not preserve the object global structure entirely \cite{lavoro}. 

Moreover, these metrics usually do not consider the spatial context of errors, treating misclassifications near the object boundaries in the same way as those in less relevant background areas. These limitations are particularly problematic in applications where spatial consistency and the integrity of segmented objects are important. Additionally, conventional metrics may be ineffective in penalizing disconnected predictions or topologically inconsistent segmentations. For instance, a prediction that splits a continuous object into several parts could receive a score comparable to a topologically correct mask, if the overall pixel-wise accuracy is high. This can lead to an segmentation quality being overestimated in cases where preserving object connectivity is essential. 

To address these issues, we propose a \textbf{Jordan-segmentatable mask}, a topology-aware segmentation mask concepts based on the Jordan Curve Theorem in the digital plane \cite{Rosenfeld_Digital_topology}. This definition uses concepts from algebraic topology, specifically homology theory \cite{homology_theory_book}, to distinguish between foreground and background in a principled and unsupervised way. To the best of our knowledge, this is the first approaches to explicitly integrate digital Jordan theory with homological invariants for image segmentation. The Jordan theorem is particularly useful in contexts where it is necessary to understand how a space can be partitioned into disjoint regions. In its simplest form, it states that a simple closed planar curve separates the plane into two connected components, one bounded (the interior) and one unbounded (the exterior). Furthermore, the Jordan theorem and its extensions play a significant role in various fields of mathematics, ranging from graph theory~\cite{MoharThomassen} to differential geometry~\cite{Bredon1993}, where they are employed to demonstrate that a given submanifold “disconnects’’ the ambient manifold into two parts.

%where they are employed to demonstrate that a given submanifold (a curve, in the basic case) “disconnects’’ the ambient manifold into two parts.

Based on this principle, we interpret segmentation masks as point sets in the digital plane and analyze their homological properties, such as the number of connected components (0-dimensional homology) and the presence of holes (1-dimensional homology).This topological approach provides a definition of segmentation that transcend spatial-level agreement by offering sensitivity to structural inconsistencies and topological distortions that conventional metrics may overlook. Our method infers topological invariants directly from the segmentation structure in an unsupervised manner, in accordance with the digital Jordan Curve Theorem. It provides a mathematically grounded and complementary tool for evaluating segmentation quality, particularly in domains where preserving topological correctness is as important as the accuracy measured by conventional pixel-, region-, or boundary-based metrics.

% The paper is divided into two main parts. Section \ref{sec:background} presents the mathematical and theoretical background, encompassing a review of the classical Jordan Curve Theorem and its digital counterpart, the essential notions from homology theory.
% Section \ref{sec:jordan_segmentable_mask} introduces the formal definition of the proposed Jordan-segmentable mask and focuses on the implementation, describing the computational pipeline resulting from our formal definition and the presentation of experimental results. \textcolor{red}{Mettere la sezione di conclusione}

The paper is organized into two main parts. Section \ref{sec:background} provides the mathematical and theoretical background, including a review of the classical Jordan Curve Theorem and its digital counterpart, as well as the key concepts from homology theory. Section \ref{sec:jordan_segmentable_mask} introduces the formal definition of the proposed Jordan-segmentable mask and details its implementation. It describes the computational pipeline derived from our formulation and presents the experimental results. Finally Section \ref{sec:conclusion} summarizes the main contributions of the work and outlines possible directions for future research.

\section{Background and basic topological notions} \label{sec:background}
This section presents the theoretical framework underlying the defined notion Jordan-segmentable mask, introducing the mathematical and topological concepts that constitute its foundation. Subsection \ref{subsec:Digital Topology} provides a formal overview of digital topology, including definitions of digital curves and connectivity. Subsection \ref{sec:jordan_theorem} reviews the classical Jordan Curve Theorem and its adaptation to the digital plane and 
subsection \ref{sec:Homology Groups and Betti Numbers} introduces fundamental concepts from algebraic topology, with a focus 
on homology groups and Betti numbers. 

\subsection{Digital Topology} \label{subsec:Digital Topology}

A \textbf{$d$-dimensional grayscale digital image} of size $(n_1,n_2,\ldots,n_d)$ is a real-valued function $ \mathcal{I}: I \rightarrow \mathbb{R}$ defined on a $d$-dimensional rectangular grid $I:=[1,n_1] \times [1,n_2] \times \ldots \times [1,n_d] \subset \mathbb{Z}^d$.
 An element $p\in I$ is called \textbf{pixel} if $d=2$, \textbf{voxel} if $d \geq 3$, and if the value $\mathcal{I}(p) \in \mathbb{R}$ the image is called \textbf{grayscale image}, else if $\mathcal{I}(p) \in \{0,1\}$ it is called \textbf{binary image}. The most commonly used images in the application are the grayscale ones, but since the geometrical properties do not depend on grayscale values, the definition of \textbf{digital topology} occurs in the case of binary images. 
 
As the definition of a binary image suggests, we can think of it as a set of points on the space—a sort of lattice— that are activated depending on on whether the image value at a specific pixel is 1 or 0. Essentially, the \textbf{digital \(d\)-space} \(\mathbb{Z}^d\) is the set of \(d\)-tuples \( x = (x_1, x_2, \dots, x_d) \) of the real Euclidean \(d\)-space with integer coordinates. The rectangular grid $I$ that defines an image, and the set of active points (those with binary value $1$) represent a subset of this space.

When working with a set of points in space, the most immediate approach is to establish proximity or adjacency relationships among them. Depending on the dimension of the digital space, different neighborhoods can be defined. We will focus specifically on the \textbf{digital plane} $\mathbb{Z}^2$, and provide a definition of adjacency and connectivity for it. 
%We will focus specifically on the \textbf{digital plane} $\mathbb{Z}^2 $, and provide for it a definition of adjacency and connectness.

\begin{definition} \label{def:adjacency}
    Given a point/pixel \( P = (x,y) \in \mathbb{Z}^2 \):  
\begin{itemize}  
    \item The \textbf{4-neighbors} of \( P \) are the four points in the horizontal and vertical directions, given by \( (x+1, y) \), \( (x-1, y) \), \( (x, y+1) \), and \( (x, y-1) \).  
    \item The \textbf{8-neighbors} of \( P \) include its 4-neighbors along with the four diagonal points: \( (x+1, y+1) \), \( (x-1, y-1) \), \( (x+1, y-1) \), and \( (x-1, y+1) \).  
    \item The \textbf{6-neighbors} of \( P \) consist of its 4-neighbors plus two diagonal points, which can be either \( (x+1, y+1) \) and \( (x-1, y-1) \) or \( (x+1, y-1) \) and \( (x-1, y+1) \).  
\end{itemize}  
Points in the 4-(8- or 6-)neighbors are called \textbf{4-(8- or 6-)adjacent}.
If \( P = (x, y) \) is a boundary point of \( I \), meaning that \( x = 1 \) or \( x = n_1 \), or \( y = 1 \) or \( y = n_2 \), we assume that the neighboring region is well-defined by considering only the portion that lies within the domain of \( I \).
\end{definition}
\begin{figure}[t]
    \centering
    \includegraphics[width=1\linewidth]{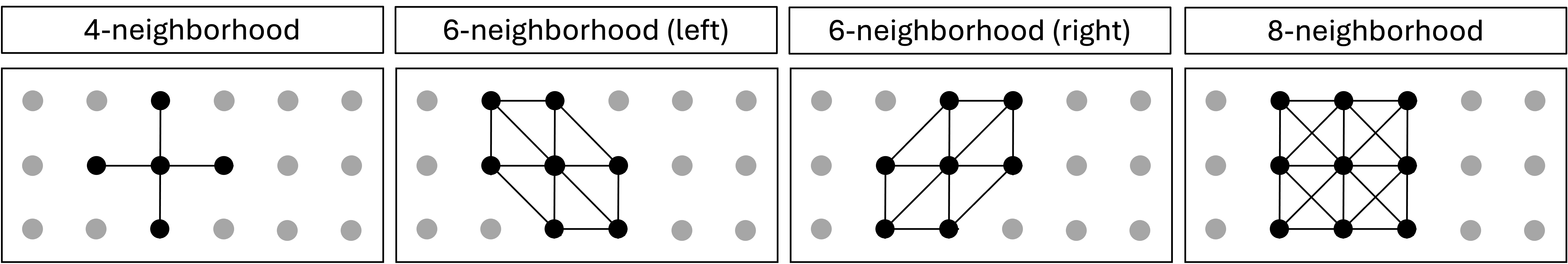}
    \caption{Illustration of the various types of neighbors for a point, along with the graph structure defined by their adjacency relations.}
    \label{fig:adjacency}
\end{figure}

The adjacency relation introduced in Definition~\ref{def:adjacency} induces a graph structure  on the points of the digital plane, where two points are connected by an edge if and only if  they are mutually adjacent, as illustrated in Figure~\ref{fig:adjacency}. 
Accordingly, we refer to \textbf{4-, 8-, or 6-graphs} to denote the graphs obtained by using  the 4-, 8-, or 6-adjacency relations, respectively.
Thus, connectivity between points is established analogously to graph theory: two points \( P \) and \( Q \) in a subset $S$ of the digital plane are said to be \textbf{4-, 8-, or 6-connected} if there exists a path between them, i.e., a sequence of distinct edges connecting a sequence of distinct vertices \cite{graph_theory}.

Moreover, a \textbf{connected component} of a set $S \subset \mathbb{Z}^2$ is a maximal subset of $S$ which is connected in the sense of previous definition. So, a topology can be obtained on $\mathbb{Z}^2$ (ad thus on $I$) using this definition of connected components, and this topology is called \textbf{Digital Topology} \cite{spazio_topologico_4_adj,spazio_topologico_6_adj,spazio_topologico_8_adj}.

\subsection{A Digital Analog of the Jordan Curve Theorem} \label{sec:jordan_theorem}

One of the most significant results in the study of the natural topology of \( \mathbb{R}^2 \) is the validity of the Jordan Curve Theorem in both directions \cite{Fary_converse_jordan_on_R^2}. A \textbf{Jordan curve}, or a \textbf{simple closed curve}, in the real plane is the image of an injective continuous map of a circle \( S^1 = \{ (x, y) \in \mathbb{R}^2 \mid x^2 + y^2 = 1 \} \) into the plane $\varphi: S^1 \to \mathbb{R}^2$. Thus, a Jordan curve \( C \) is a continuous loop that has no self-intersections.
The Jordan Curve Theorem asserts that any simple closed curve divides the plane into two distinct regions: the interior, which is bounded, and the exterior, which is unbounded. It can be stated as follows. %Formally, the theorem can be stated as follows:
\begin{theorem}\textbf{(Jordan Curve Theorem on \( \mathbb{R}^2 \))} \label{thm:Jordan_real}  \\
Let \( C \) be a Jordan curve in the plane \( \mathbb{R}^2 \). Then its complement, \( \mathbb{R}^2 \setminus C \), consists of exactly two connected components. One of these components is bounded (referred to as the \textit{interior}), while the other is unbounded (referred to as the \textit{exterior}).
\end{theorem}

\begin{figure}[t]
    \centering
    \includegraphics[width=0.9\linewidth]{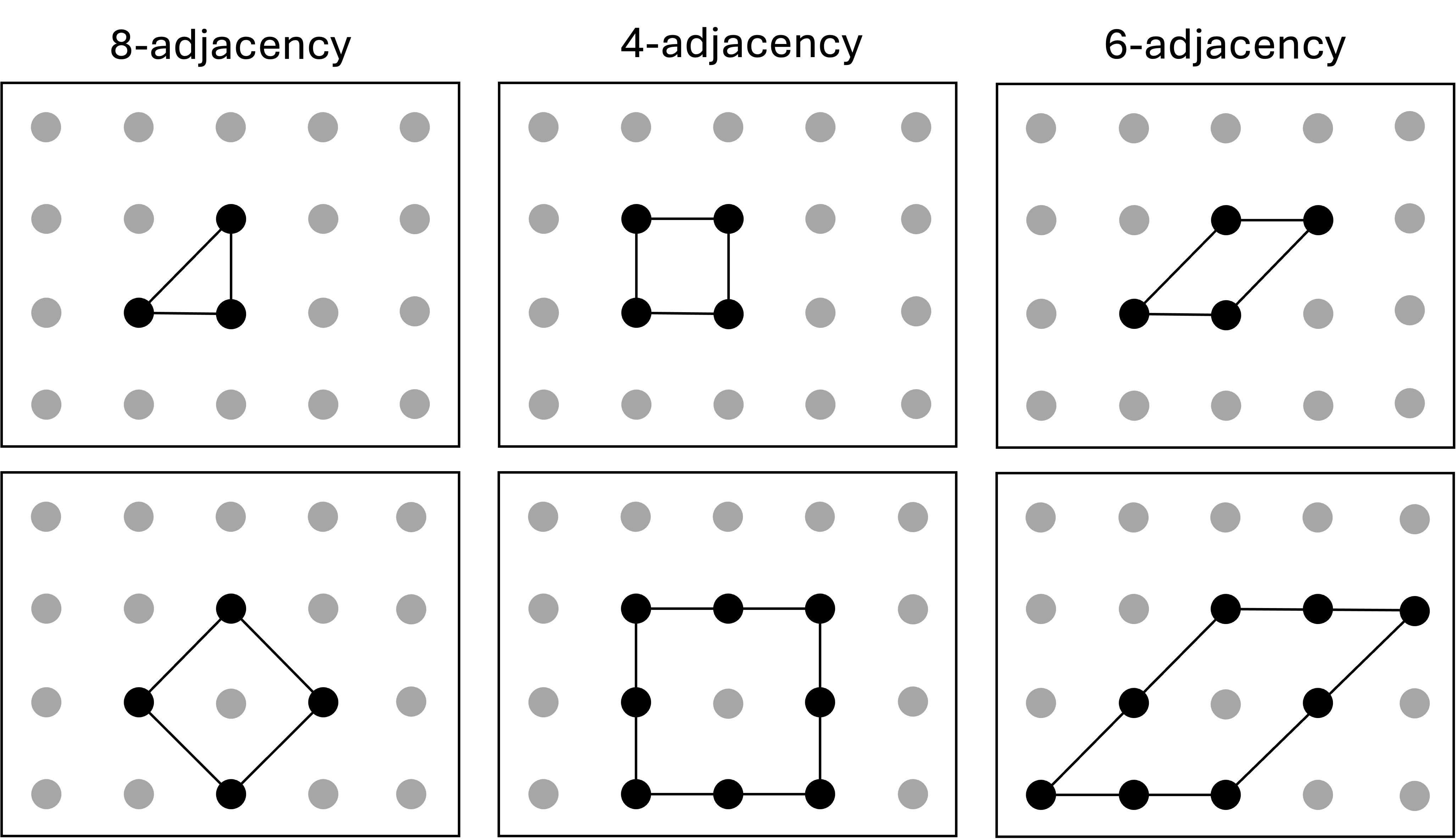}
    \caption{Illustration of the minimal number of points required to define a digital curve.}
    \label{fig:problema_minimi_pti}
\end{figure}

\begin{figure}[t]
    \centering
    \includegraphics[width=0.6\linewidth]{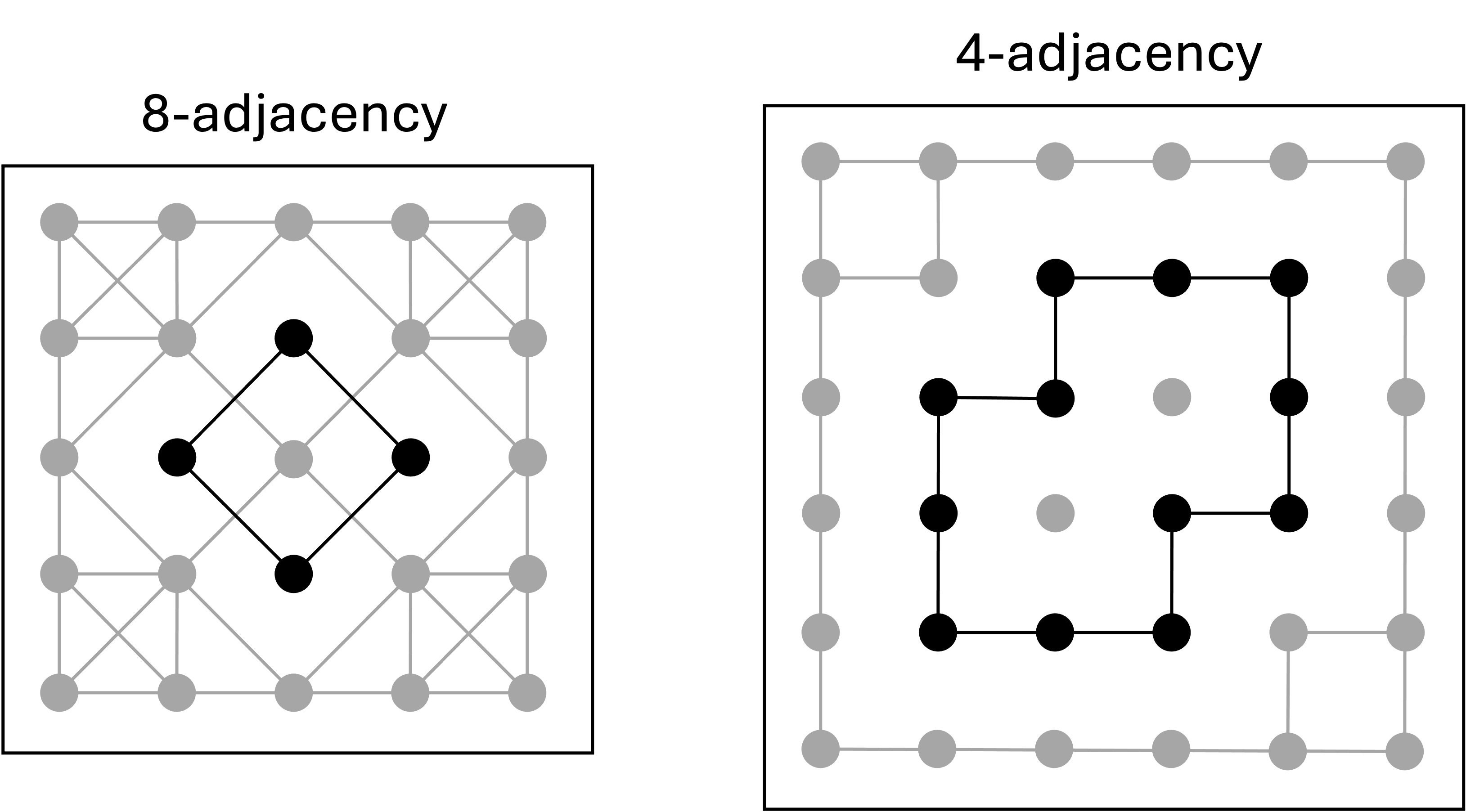}
    \caption{Examples illustrating connectivity paradoxes in the digital plane, for 8- and 4- adjacency over $S$ and its complementary.}
    \label{fig:connectivity_paradoxa}
\end{figure}

To establish an analogue of the Jordan Curve Theorem in the digital plane, based on Theorem \ref{thm:Jordan_real}, it is necessary to clarify some preliminary concepts.
The first fundamental question concerns the definition of a curve in the digital plane \( \mathbb{Z}^2 \). Following Rosenfeld's formulation in \cite{Rosenfeld_Digital_topology}, a subset \( S \subset \mathbb{Z}^2 \) is called a \textbf{4-, 6-, or 8-curve} if it is connected and each of its points has exactly two 4-, 6-, or 8-adjacent neighbors within \( S \), respectively.  
As the aim is to formulate a Jordan theorem for the digital plane, degenerate cases are excluded by assuming that a 4-curve and a 6-curve must contain at least eight points, while an 8-curve must contain at least four points. As illustrated in Figure~\ref{fig:problema_minimi_pti}, the presence of only three or four points in these degenerate cases is insufficient to ensure the existence of even a single interior point. 
It can be shown that a digital curve can enclose at most one hole; however, this does not necessarily hold when using the same adjacency (4- or 8-) for both \( S \) and its complement. This issue is illustrated Figure~\ref{fig:connectivity_paradoxa}. In the first example, the set of points forms an 8-curve but does not qualify as a 4-curve, as its complement consists of a single 8-connected component. Conversely, the second example represents a 4-curve but not an 8-curve, with its complement comprising three distinct 4-connected components. This phenomenon is commonly referred to as the \textit{connectivity paradox}.
Provided that different adjacency relations must be used for \( S \) and its complement, Rosenfeld in \cite{Rosenfeld_Arcs_and_Curves} proved that a digital curve encloses exactly one hole, thereby establishing the validity of the Jordan Curve Theorem in the digital setting. This result can be formally stated as follows:
\begin{theorem}\textbf{(Digital Jordan Curve Theorem)} \label{theorem:Jordan theorem on R^2}  \\
Let \( S \) be a simple 4-(6- or 8-)curve in the digital plane \( \mathbb{Z}^2 \) containing at least eight (or four) points. Then its complement, \( \mathbb{Z}^2 \setminus S \), consists of exactly two connected components with respect to 8-(6-, or 4-)adjacency, and every point in \( S \) is 8-(6-, or 4-)adjacent to at least one point in both of these components.
\end{theorem}
\noindent Additionally, Rosenfeld proved the converse of the Jordan Curve Theorem for the digital plane in \cite{ROSENFELD_converse_jordan}:
\begin{theorem}\textbf{(Converse of the Digital Jordan Curve Theorem)} \label{theorem: Converse digital jordan}\\
Let \( S \) be a subset of the digital plane \( \mathbb{Z}^2 \) satisfying the following conditions:
\begin{itemize}
    \item The complement \( \mathbb{Z}^2 \setminus S \) consists of exactly two connected components, \( D \) and \( E \), with respect to 8-, 4-, or 6-adjacency.
    \item Every point in \( S \) is 8-, 4-, or 6-adjacent to at least one point in both \( D \) and \( E \).
\end{itemize}
Then \( S \) forms a simple 4-, 8-, or 6-curve, containing at least four or eight points.
\end{theorem}
\noindent When considering the theorems for a subset of the digital plane (i.e., the image \(I\)), it is assumed that \(S\) is never located on the image boundaries. This assumption prevents degenerate cases.

\subsection{Homology Groups and Betti Numbers}
\label{sec:Homology Groups and Betti Numbers}
A fundamental approach to identifying whether an object in a topological space corresponds to a connected component or a loop is through simplicial homology groups.
Defining a graph structure on the image based on various adjacency relations between points, results in a simplicial complex structure \cite{simplicial_complex_ref} on it, using as simplices active (or non-active) points, edges between them, and triangles identified in the graph. Formally, given \(V\) a finite non-empty set, an \textbf{abstract simplicial complex} \(K\) on \(V\) is a collection of non-empty subsets of \(V\) such that, for any \(\sigma \in K\), every non-empty subset \(\tau \subset \sigma\) is also in \(K\). So a graph is an abstract simplicial complex.

A \textbf{k-chain} is a (finite) linear combination of \( k \)-simplices, and the collection of all \( k \)-chains within a simplicial complex \( K \) is denoted as \( C_k \). When endowed with an addition operation, this set forms a group, known as the \textbf{chain group}, upon which algebraic operations can be defined.
The first fundamental operation we define is the \textbf{boundary homomorphism}, a map \(\delta_k: C_k \to C_{k-1}\) that acts on each \( k \)-simplex as  
\(
\delta_k([x_1, \dots, x_{k+1}]) = \sum_{i=1}^{k+1} (-1)^i [x_1, \dots, x_{i-1}, x_{i+1}, \dots, x_{k+1}],
\) 
and extends linearly to \( k \)-chains.\\
Boundary homomorphism is the map that represents a linear combination of $k$-simplexes as a linear combination of $(k-1)$-simplexes, i.e., their boundaries. This map enables the definition of two fundamental groups: \textbf{k-cycles} $Z_k=Ker(\delta_k)$ and the \textbf{k-boundaries} $B_k=Im(\delta_{k+1})$. The group $Z_k$ consists of the linear combinations of $k$-simplices whose boundary is zero, whereas $B_k$ includes all linear combinations of $k$-simplices that are boundaries of $(k+1)$-chains. Since the composition of two consecutive boundary homomorphisms satisfies \( \delta_k \circ \delta_{k+1} = 0 \) \cite{distanze_PB}, it follows that \( B_k \subseteq Z_k \), ensuring that the quotient group \( Z_k / B_k \) is well-defined.

The \textbf{\( k \)-th simplicial homology group} \cite{libro_algebraic_topology} of a simplicial complex \( K \) is given by the quotient \( H_k = Z_k / B_k \), where its elements are known as \textbf{homology classes}. 
Although homology groups are defined in an abstract way, they capture fundamental the topological properties of the underlying topological space. 
In particular, Betti numbers can be used to numerically evaluate \textbf{topological invariants} — objects in a simplicial complex with a specific topological property (e.g. a connected component or a loop) — of a specific topological space under analysis.

By definition, the $\textbf{k-th Betti number}$ associated to the $k$-simplicial homology group $H_k$ of a generic space $G$ is $\beta_k=rk(H_k(M))$
and counts the number of linearly independent topological invariants in $G$. Thus, if $G$ is a graph, $\beta_0(G)$ counts the number of connected components of it and $\beta_1(G)$ counts the number of loops; so, a Jordan curve \( S \) will have \( \beta_0(S) = 1 \) and \( \beta_1(S) = 1 \). Moreover, by Jordan thorem, it will partition the digital plane (or the real plane) into two connected components, meaning that the complement of \( S \) will satisfy \( \beta_0 = 2 \).

\section{Jordan-Segmentatable mask} \label{sec:jordan_segmentable_mask}
In order to introduce the concept of topology-aware segmentation, we must first formalize the process by which a digital image is partitioned into foreground and background regions. We begin by defining binary segmentation, on which the subsequent definition of a Jordan-segmentable mask is based. A Jordan-segmentable mask is a segmentation that satisfies the topological separation property implied by the Jordan theorem.

\begin{definition} \textbf{(Binary Segmentation)} \\
Let $\mathcal{I}: I \rightarrow \mathbb{R}$ be a $d$-dimensional grayscale digital image 
of size $(n_1, n_2, \ldots, n_d)$, where 
$I := [1, n_1] \times [1, n_2] \times \ldots \times [1, n_d] \subset \mathbb{Z}^d$ 
denotes the discrete image domain.  
A \textbf{binary segmentation} of $\mathcal{I}$ is an image
$$
\mathcal{I}_s : I \rightarrow \{0,1\},
$$
s.t. $\mathcal{I}_s^{-1}(0)=I_0$ and $\mathcal{I}_s^{-1}(1)=I_1$ where $I_0, I_1 \subset I$, $I = I_0 \cup I_1$, $I_0 \cap I_1 = \emptyset.$
\end{definition}

\noindent We define the \textbf{segmentation mask} as 
\(
M := \{\, p \in I \mid \mathcal{I}_s(p) = 1 \,\} = I_1,
\)
i.e., the set of active points in the digital domain where the segmentation takes value \(1\).

Ideally, this set represents the region corresponding to the object of interest, although this correspondence is not guaranteed. This definition does not ensure a proper partition of the image domain into an ``inside'' and an ``outside'' region, as required by the Jordan Curve Theorem. Consequently, no topological separation property is generally implied. 
Our goal is therefore to introduce the concept of a Jordan-segmentable mask to ensure a spatial partition of the digital domain that is consistent with the Jordan theorem. 
In this sense, the segmentation should guarantee the existence of a subset that divides the digital space into two disjoint, complementary topological regions.

The conditions required for the application of the digital Jordan theorem are not inherently satisfied given the structure of the segmentation mask $M$. Therefore it is necessary to identify a subset $S \subseteq I$ that explicitly fulfills these conditions, enabling the theorem to be applied effectively to partition the image domain into two distinct regions.

\begin{figure}[t]
\centering  
\includegraphics[width=0.9\linewidth]{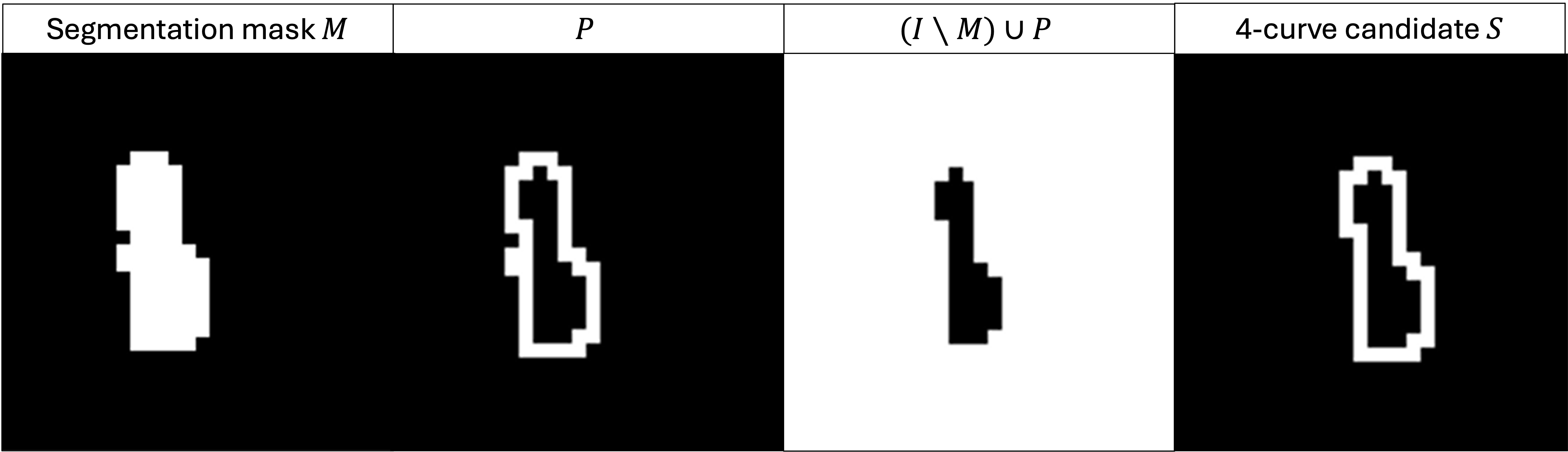}  
\caption{Construction of the 4-curve candidate $S$.}  
\label{fig:boundary_construction}  
\end{figure}  

\begin{definition}\textbf{(4-curve candidate)}\\
Let \( \mathcal{I}_s \) be a binary segmentation of a digital image, and let \( M \) denote its segmentation mask. 
We define the \textbf{4-curve candidate} \( S \subseteq M \) as the subset obtained through the following procedure:
\begin{enumerate}
    \item Identify the set \( P \subseteq M \) of points that are 8-adjacent to at least one point with value 0. 
    This yields a preliminary boundary set that is 8-adjacent to one of the two regions in Theorem~\ref{theorem: Converse digital jordan}, but not necessarily to both.
    \item Form the union of \( P \) with the complement of the segmentation mask, i.e.\ \( (I \setminus M) \cup P \).
    \item Within this union, extract the points with value 1 that are 8-adjacent to at least one point with value 0. 
    The resulting set is denoted by \( S \) and referred to as the \textbf{4-curve candidate}.
\end{enumerate}
\end{definition}
\noindent Intuitively, the set \( S \) represents a sort of boundary structure that satisfies the adjacency conditions required by the digital Jordan theorem and can therefore be tested for being a valid $4$-curve. The construction steps are shown in Figure \ref{fig:boundary_construction}.

Demonstrating that the set \(S\) forms a $4$-curve, contains at least eight points and partitions the plane into two connected components (an interior region and an exterior region), provides a validation of the Jordan theorem in both directions. 
To prove the first part of the theorem, we consider the Betti numbers of the $4$-adjacency graph associated with the $4$-curve. As this graph forms a simplicial complex, the computed values of the Betti numbers, namely \(\beta_0 = 1\) and \(\beta_1 = 1\) characterize the set \( S \) as a $4$-curve, as described in subsection~\ref{sec:Homology Groups and Betti Numbers}.

\begin{figure}[thb]
\centering
\includegraphics[width=1\linewidth]{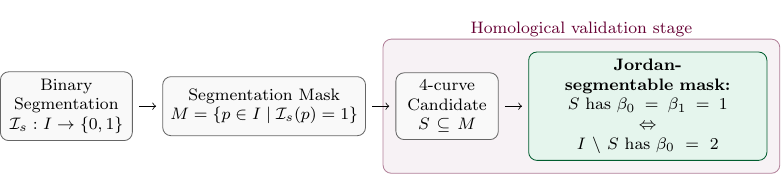}

\caption{Conceptual workflow illustrating the proposed method for an homological validation of a  Jordan-segmentable mask from binary segmentation.}
\label{fig:jordan_pipeline_refined}
\end{figure}

The second implication of the theorem can be verified by considering the 8-adjacency graph on the complement of \(S\). The outcome \(\beta_0 = 2\), confirms that the image is partitioned into exactly two distinct connected components, thereby validating the theorem.
The topological conditions expressed through the Betti numbers provide a rigorous framework that ensures the segmentation satisfies the spatial separation property implied by the Jordan theorem.
\begin{definition}\textbf{(Jordan-segmentable mask)}\\
A segmentation mask $M$ is said to be \textbf{Jordan-segmentable} 
if there exists a 4-curve candidate $S \subseteq M$ that forms a 4-curve in the digital sense, i.e.,\ whose 4-adjacency graph satisfies $\beta_0 = \beta_1 = 1$, 
or equivalently whose complement $I \setminus S$ is 8-segmented into two components ($\beta_0=2$).
\end{definition}
Figure \ref{fig:jordan_pipeline_refined} provides a visual and formal summary of the proposed method. Starting from a binary segmentation mask, a $4$-curve candidate is identified and validated using topological conditions (Betti numbers). 
If these conditions are met, the mask can be conclusively considered Jordan-segmentable, ensuring a proper topological separation of the digital space into interior (foreground) and exterior (background) regions.

\subsection{Jordan-segmentable mask in practice} \label{sec:implementation}

This section describes the practical implementation of the proposed method to understand if a binary segmentation is a Jordan-segmentable mask, along with illustrative results on the FSS-1000 dataset.

\begin{figure}[htb]  
\centering  
\includegraphics[width=1\linewidth]{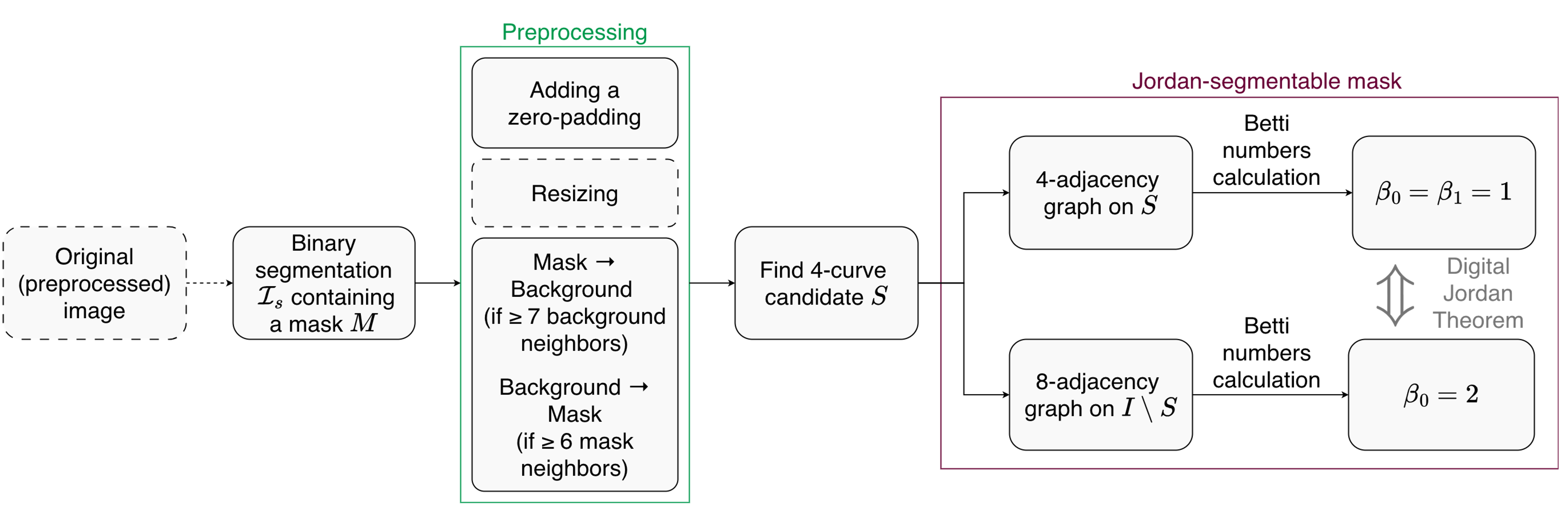}  
\caption{Pratical implementation of the proposed method.}  
\label{fig:pipeline}  
\end{figure} 

Given a binary image containing a segmentation mask $M$, our goal is to assess whether $M$ is a Jordan-segmentable mask. 
To satisfy the assumptions of the digital Jordan theorem, preprocessing of the binary image is required. In particular, since the set \(S\) is assumed not to lie on the image boundary, a zero-padding is added around the image to prevent degenerate cases.
Moreover, to eliminate small inconsistencies in the segmentation masks, pixels belonging to $M$ are converted to background pixels if surrounded by more than seven background neighbors, while background pixels are converted to mask pixels if they have more than six mask neighbors.\\
Then, the 4-curve candidate $S \subseteq M$ is extracted according to the procedure described previously. 
The candidate $S$ represents a boundary structure that can potentially satisfy the adjacency conditions of the digital Jordan theorem. 

Demonstrating that $S$ forms a 4-curve containing at least eight points and partitions the image domain into exactly two connected components (interior and exterior) establishes that $M$ is Jordan-segmentable. 
To validate this formally, a 4-adjacency graph is constructed on $S$ and its Betti numbers are computed. 
If $\beta_0 = 1$ and $\beta_1 = 1$, the set $S$ is confirmed as a 4-curve. 
The complement $I \setminus S$ is then analyzed via its 8-adjacency graph; obtaining $\beta_0 = 2$ confirms that the domain is partitioned into exactly two connected components, thus satisfying the topological conditions required by the theorem. 
Figure~\ref{fig:pipeline} provides a visual summary of the pipeline: starting from a binary segmentation mask, a 4-curve candidate is identified and validated using topological criteria, resulting in a Jordan-segmentable mask when all conditions are satisfied.

\begin{figure}[tb]
    \centering
    \includegraphics[width=1\linewidth]{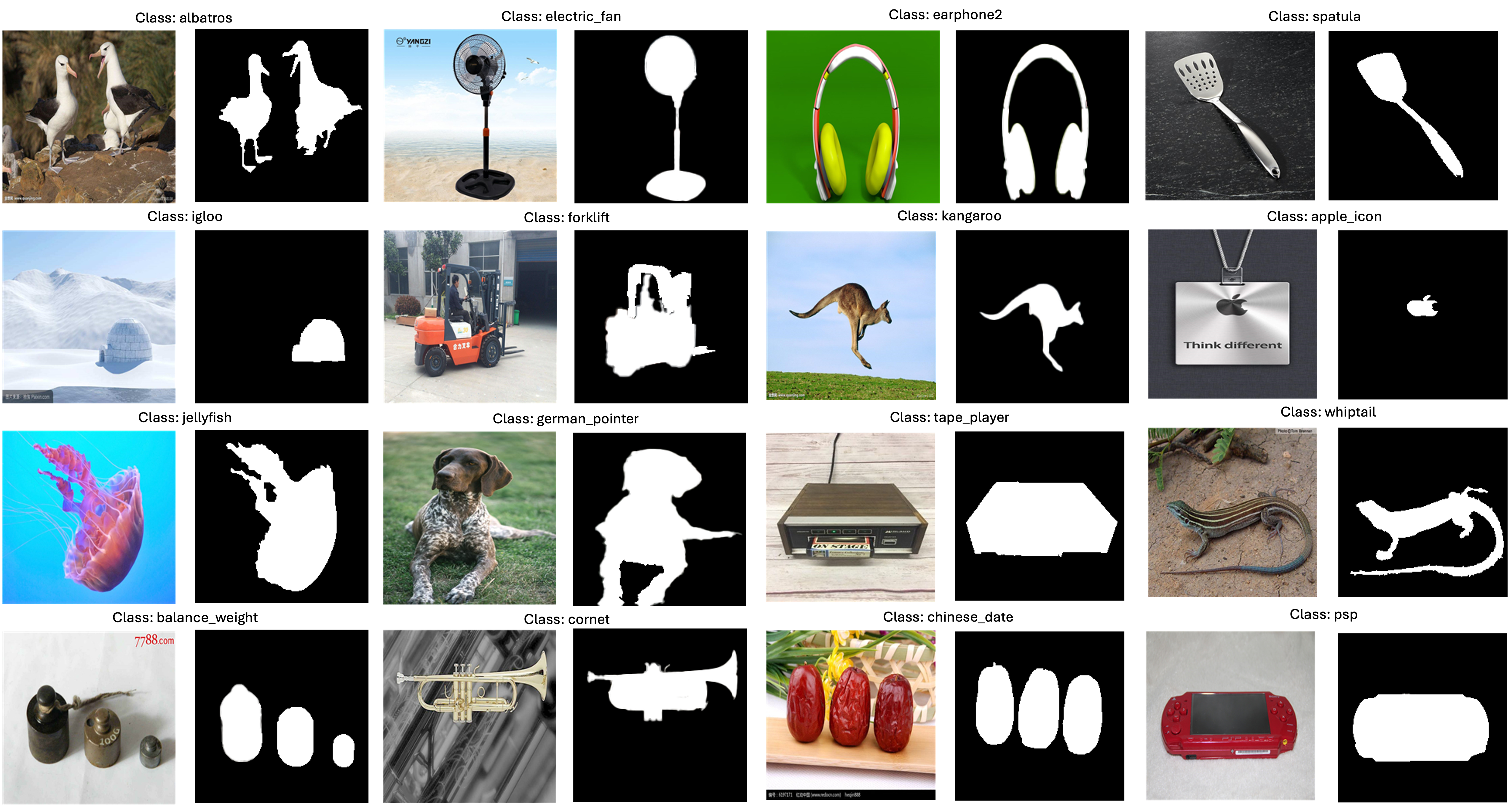}
    \caption{Some images from the FSS-1000 dataset, along with their corresponding classes and segmentation annotated masks.}
    \label{fig:dataset}
\end{figure}

To evaluate the proposed definition of a Jordan-segmentable mask computationally, we conduct experiments on the FSS-1000 dataset \cite{dataset_FSS-1000}, considering both the original annotation masks provided with the dataset and those obtained by applying dedicated segmentation methods to the original images.
The FSS-1000 dataset is designed for few-shot object segmentation and is large-scale. It contains 1,000 object classes, each with 10 images of dimension $144 \times 144$ and pixel-wise segmentation annotations. Some examples are shown in
Figure \ref{fig:dataset}, along with their annotations and classes. A key feature of FSS-1000 is that it includes many object classes not found in other datasets.  It is structured to emphasize a large number of object classes rather than a large number of images per class. The images were gathered from multiple internet search engines to avoid bias, and annotations were performed using Photoshop’s “quick selection” tool, that allow for semi-automatic object selection, which was then manually refined.

\begin{figure}[tb]
\centering
    \includegraphics[width=0.8\linewidth]{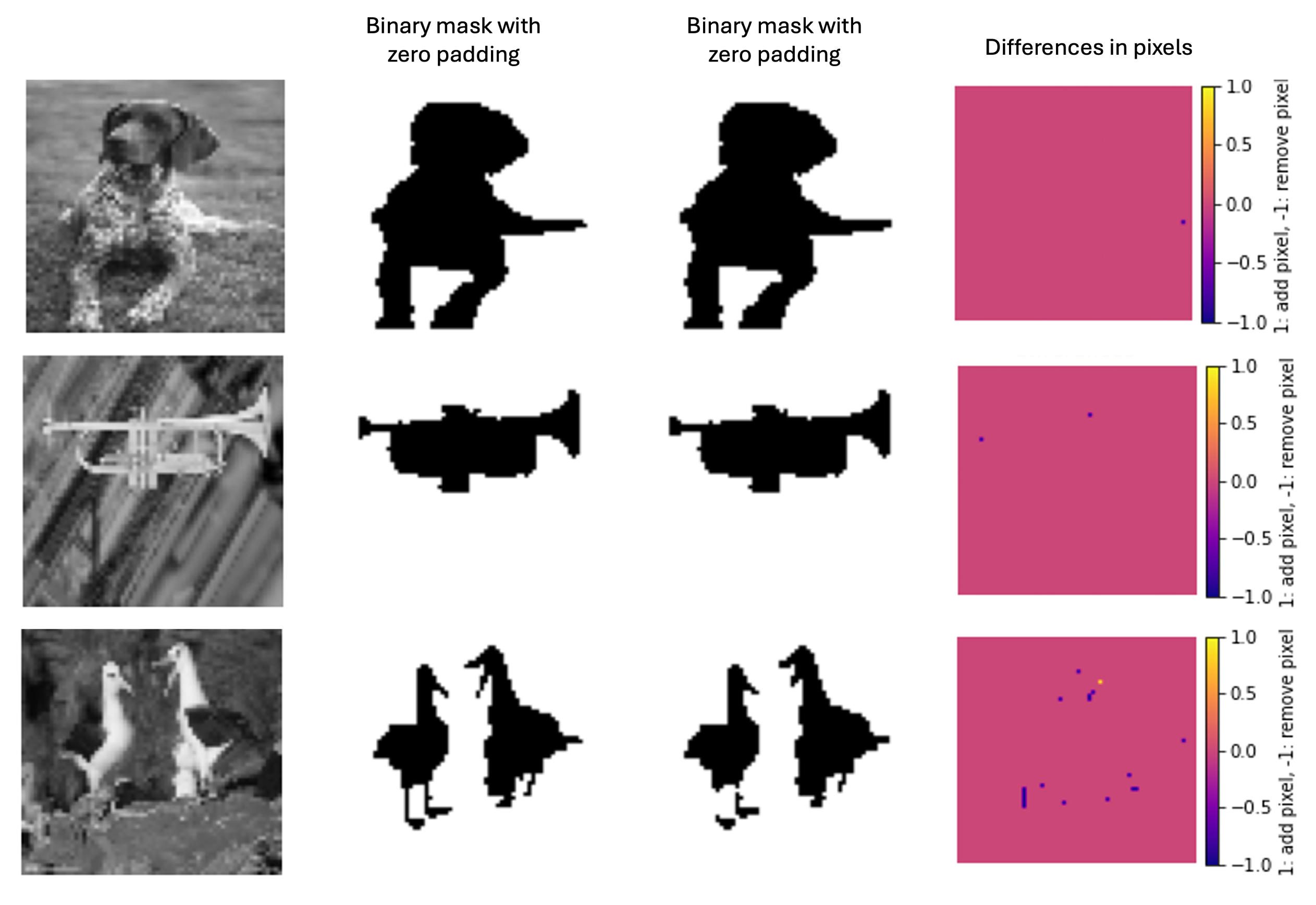}
 \caption{
Examples of preprocessing are shown. From left to right: the original image, the binarized mask with zero-padding, the preprocessed mask, and a visualization of the pixels added or removed during preprocessing, where added pixels are assigned a value of 1 and removed pixels a value of -1. The masks are displayed in black in this figure for visualization purposes. Preprocessing procedure introduces only minor changes in the first two cases, whereas in the third case it is essential for removing spurious regions that do not belong to the object of interest.}
\label{fig:preprocessing}
\end{figure}

As a preprocessing step, all annotation masks are resized to $64\times 64$ pixels to reduce computational cost. Each mask then undergoes the preprocessing procedure described above. As shown in Figure~\ref{fig:preprocessing}, this step only makes minor changes in the first two cases. However, in the third case it is essential for removing spurious regions that do not belong to the object of interest.

\begin{figure}[t!]
    \centering
    \includegraphics[width=0.6\linewidth]{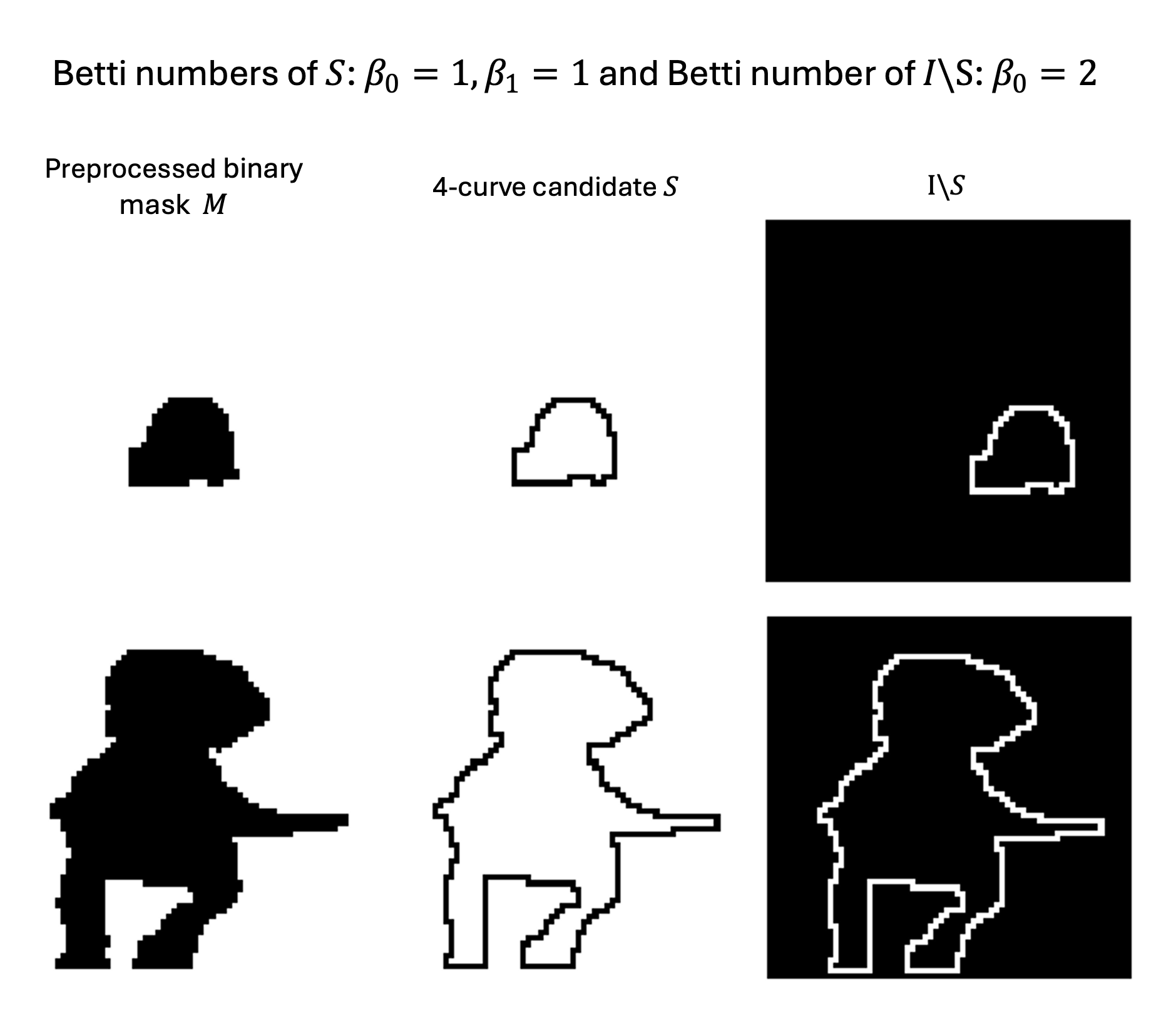}
    \caption{The segmentation mask is a Jordan-segmentable mask.}
    \label{fig:jordan verificato}
\end{figure}

\begin{figure}[t!]
    \centering
    \includegraphics[width=0.6\linewidth]{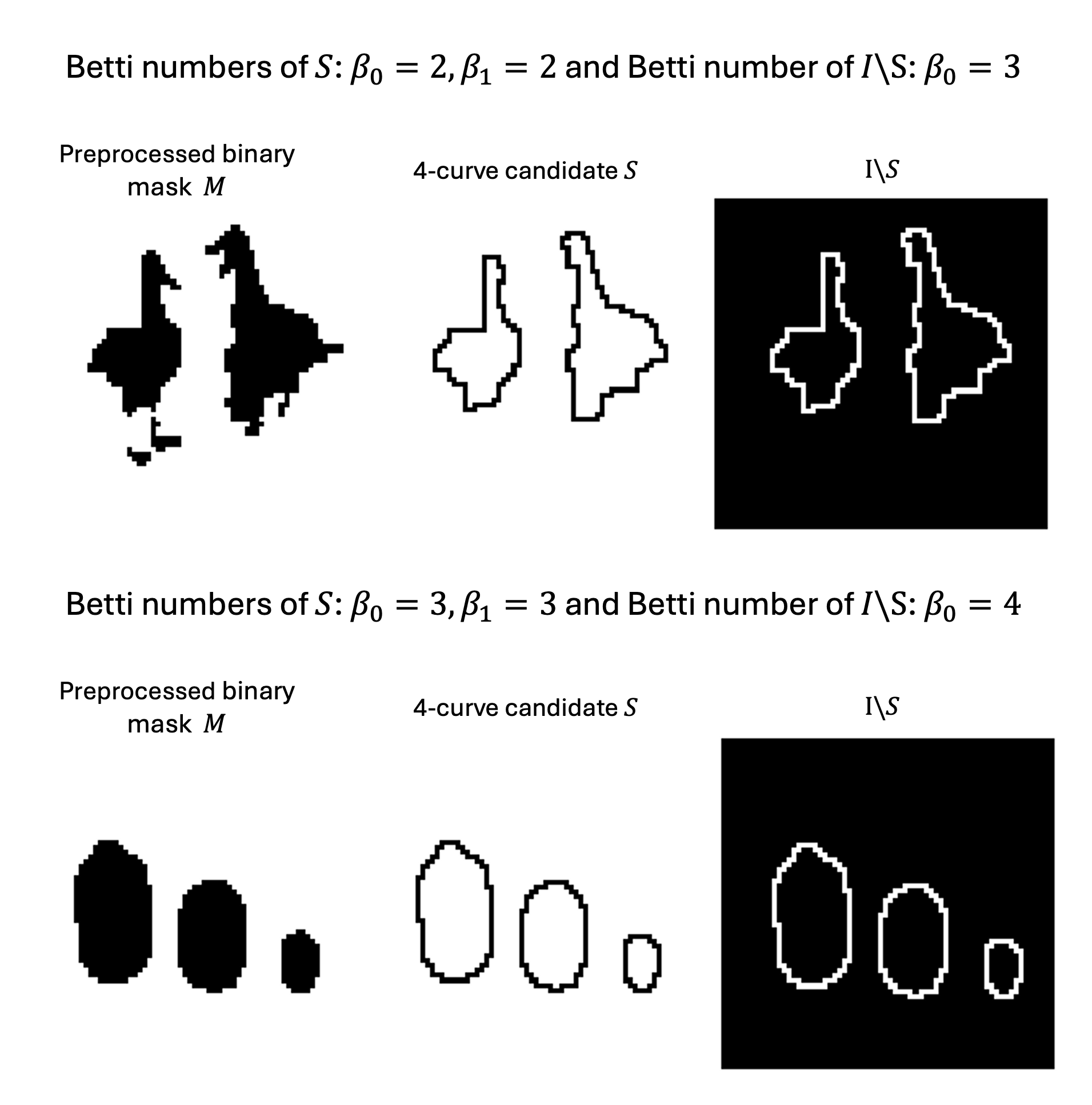}
    \caption{The segmentation mask is a Jordan-segmented mask identifying more then one object.}
    \label{fig:jordan verificato su più oggetti}
\end{figure}

\begin{figure}[t!]
    \centering
    \includegraphics[width=0.6\linewidth]{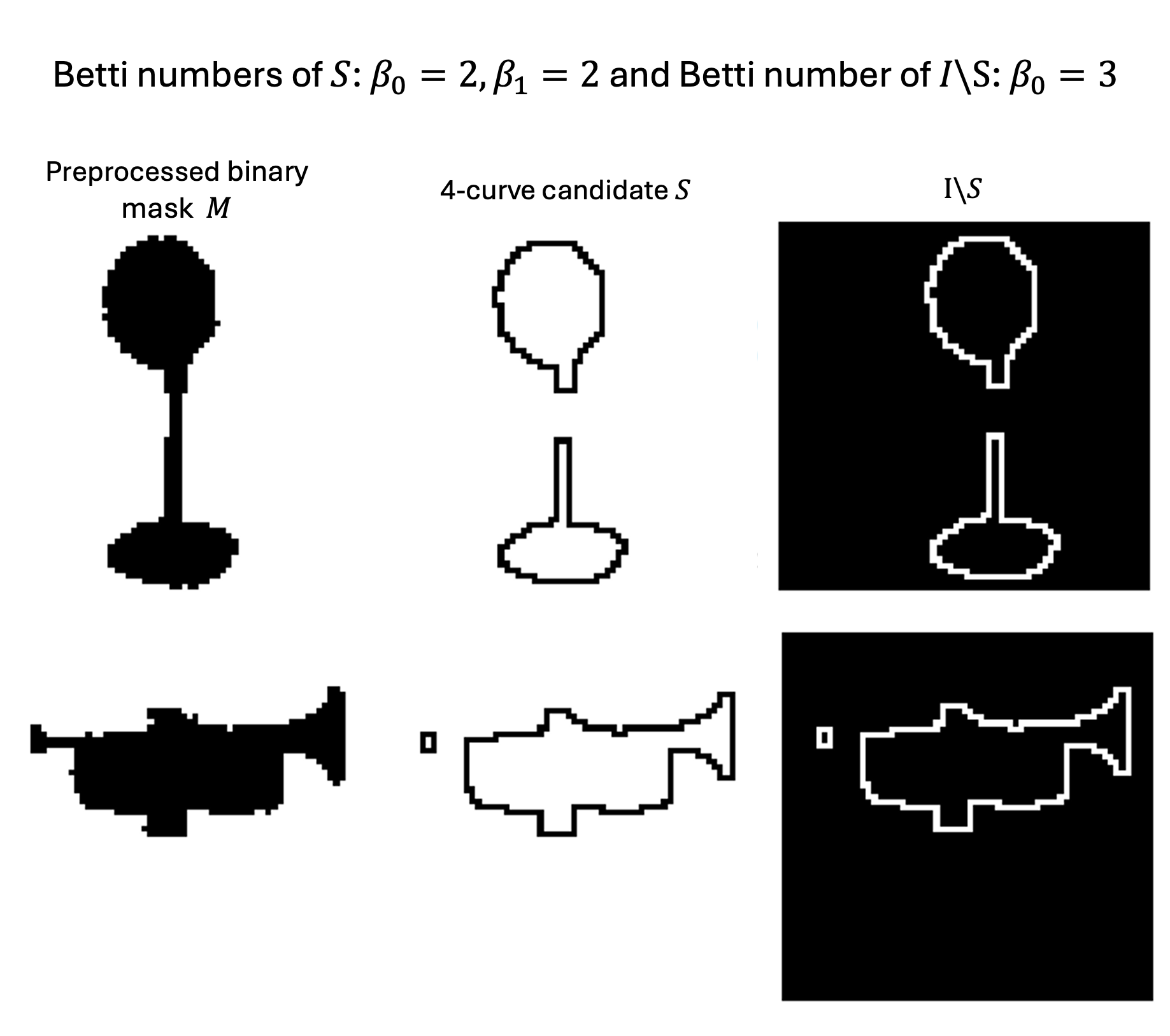}
    \caption{The segmentation mask corresponds to a Jordan-segmented mask that represents a single object fragmented into multiple connected components.}
    \label{fig:Jordan verificato spezzettando un oggetto}
\end{figure}

\begin{figure}[t!]
    \centering
    \includegraphics[width=0.6\linewidth]{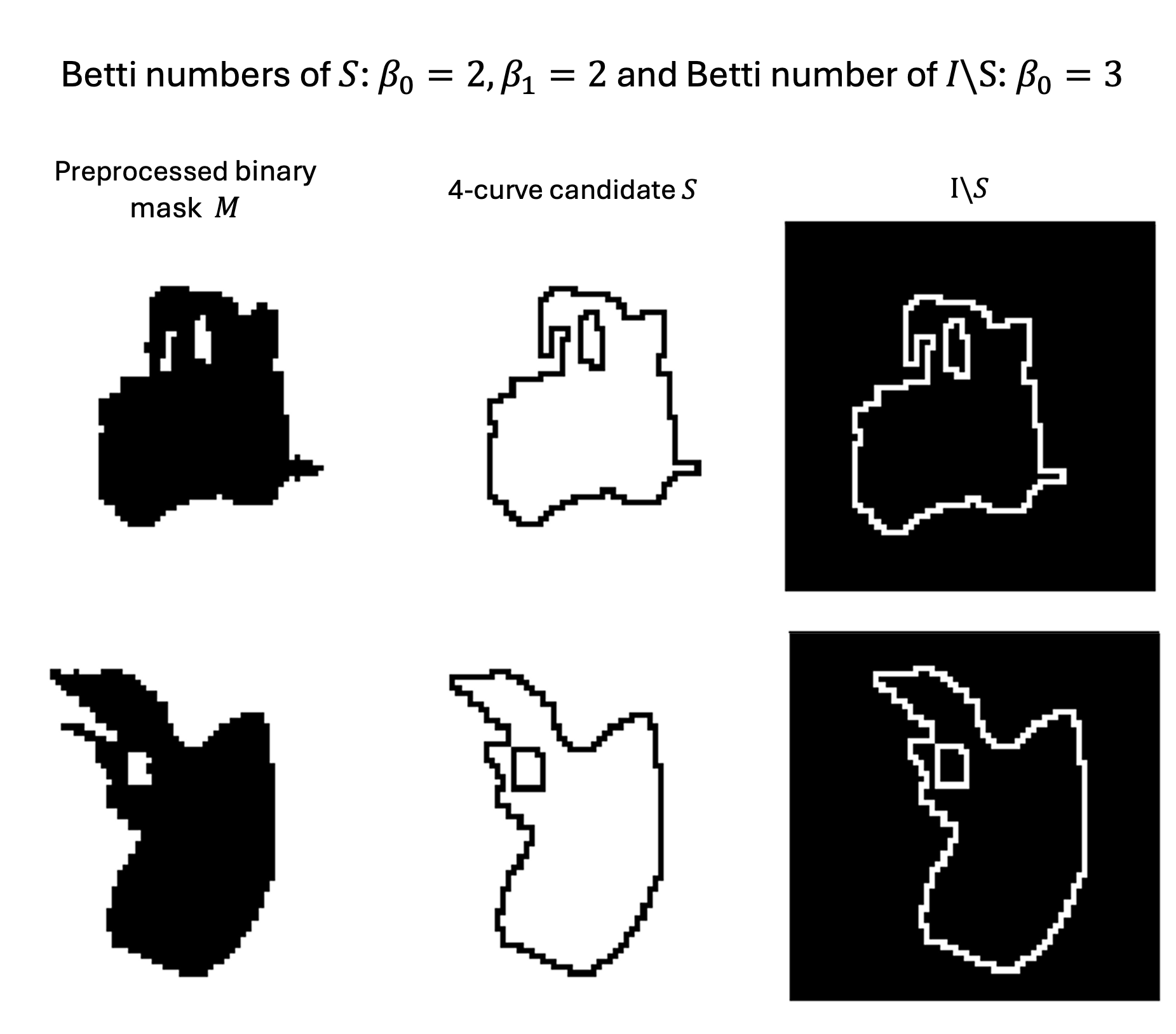}
    \caption{The segmentation mask corresponds to a Jordan-segmented mask for objects with holes.}
    \label{fig:Jordan verificato su oggetto bucato}
\end{figure}

\begin{figure}[t!]
    \centering
    \includegraphics[width=0.6\linewidth]{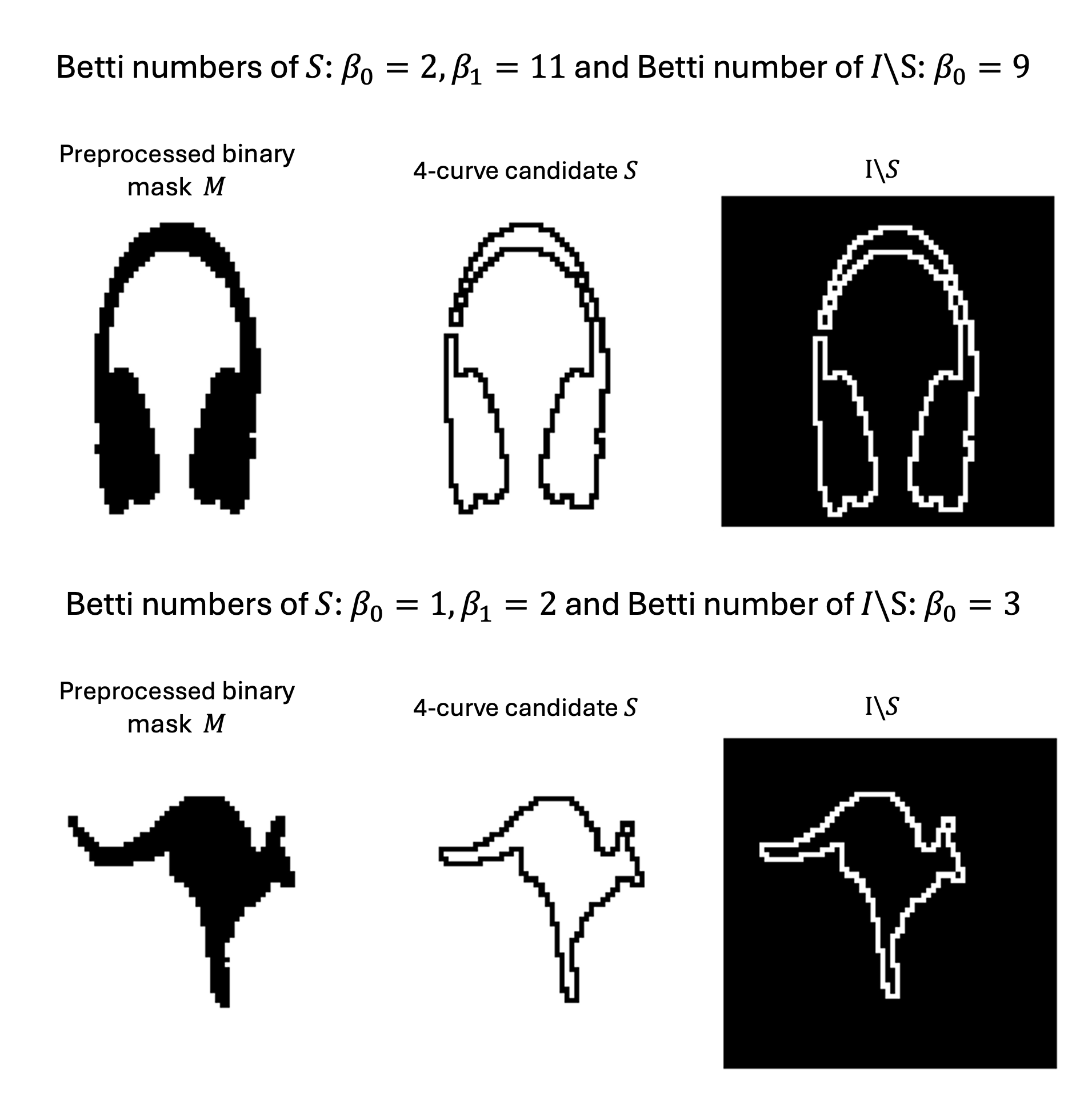}
    \caption{The segmentation mask is not a Jordan-segmented mask.}
    \label{fig:Jordan non verificato}
\end{figure}

Applying the proposed method, which leads from binary segmentation to aJordan-segmentable mask, to a selection of randomly sampled image annotations from the dataset, four distinct behaviors can be identified:
\begin{itemize} 
\item \textbf{The segmentation mask is a Jordan-segmented mask in the described sense.} As illustrated by the examples in Figure \ref{fig:jordan verificato}, the theorem holds in both directions as confirmed by Betti number calculations. 
\item \textbf{The segmentation mask is a Jordan-segmented mask that identifies more than one object.} As shown in Figure \ref{fig:jordan verificato su più oggetti}, although the Betti numbers may initially appear misleading, the theorem effectively applies to each object separately. Ultimately, the background is accounted for only once by the Betti number $\beta_0$.

\item \textbf{The segmentation mask corresponds to a Jordan-segmented mask representsing a single object fragmented into multiple connected components.} As illustrated in Figure \ref{fig:Jordan verificato spezzettando un oggetto}, the method used to obtain the 4-curve candidate $S$ may divide a single object into multiple sections. Nevertheless, the theorem is satisfied in each section, analogously to the previous case.

\item \textbf{The segmentation mask corresponds to a Jordan-segmented mask for objects with holes.} As shown in Figure \ref{fig:Jordan verificato su oggetto bucato}, objects that inherently contain holes are interpreted as if the Jordan theorem were applied to two objects: once to the object itself and once to the hole in the object. As the previous case, the background is ultimately counted only once.

\item \textbf{The segmentation mask is not a Jordan-segmented mask.} There are a few rare cases, illustrated in Figure \ref{fig:Jordan non verificato}, where the detected potential $4$-curve does not fulfil the definition of a simple curve and thus the theorem does not apply. This usually happens when thin diagonal sections are treated as sequences of separate holes, which disrupts the calculation of the first Betti number $\beta_1$ within the $4$-curve.
\end{itemize}

As discussed in Section \ref{sec:jordan_segmentable_mask}, binary segmentations consist of images partitioned into two regions. Typically, segmentation is obtained by exploiting image intensity, statistical properties, or other features to separate these two areas. To evaluate the proposed approach, the focus is on a set of widely adopted segmentation techniques: Otsu's method, the Ridler–Calvard method, k-means segmentation and the watershed algorithm. These classical methods offer diverse strategies and provide a comprehensive framework for evaluating the validity of the proposed methodology. 
Thresholding methods are among the most fundamental and widely used approaches for binary image segmentation. These techniques assume that pixel intensity values can be separated into two distinct classes --foreground (usually the object) and background -- based on a selected threshold. Specifically, \textbf{Otsu's method}~\cite{Otsu1979} determines the optimal threshold by maximizing the between-class variance of the pixel intensities, effectively enhancing the separability of the two classes. In contrast, the \textbf{Ridler-Calvard method} \cite{RidlerCalvard1978} adopts an iterative strategy. Starting from an initial estimate—typically the global mean intensity -- the algorithm partitions the image into two regions, it computes the mean intensity of each region, and updates the threshold as the average of these two means. This process is repeated until convergence is reached, resulting in a stable threshold value. Clustering-based segmentation treats pixel classification as an unsupervised learning problem, grouping pixels according to similarity in intensity or other features. \textbf{K-means} clustering is an example of this approach \cite{MacQueen1967}; it is an iterative partitioning algorithm that assigns each pixel to one of $k$ clusters by minimizing the intra-cluster variance. For binary segmentation, $k=2$ is used. The algorithm alternates between assigning pixels to the nearest cluster centroid and updating the centroids based on the mean of assigned pixels. The \textbf{watershed algorithm} \cite{BeucherLantuejoul1979} segments an image by interpreting its gradient magnitude as a topographic surface. The gradient magnitude of the grayscale image is computed, where high values correspond to edges. The algorithm then identifies its local minima, representing catchment basins. Starting from these minima, the algorithm simulates a flooding process where water gradually fills the basins. Where two flooding regions meet, a boundary, called a watershed line, is formed to separate them. These watershed lines define the segmentation contours.

\begin{figure}[tb]
    \centering
    \includegraphics[width=0.85\linewidth]{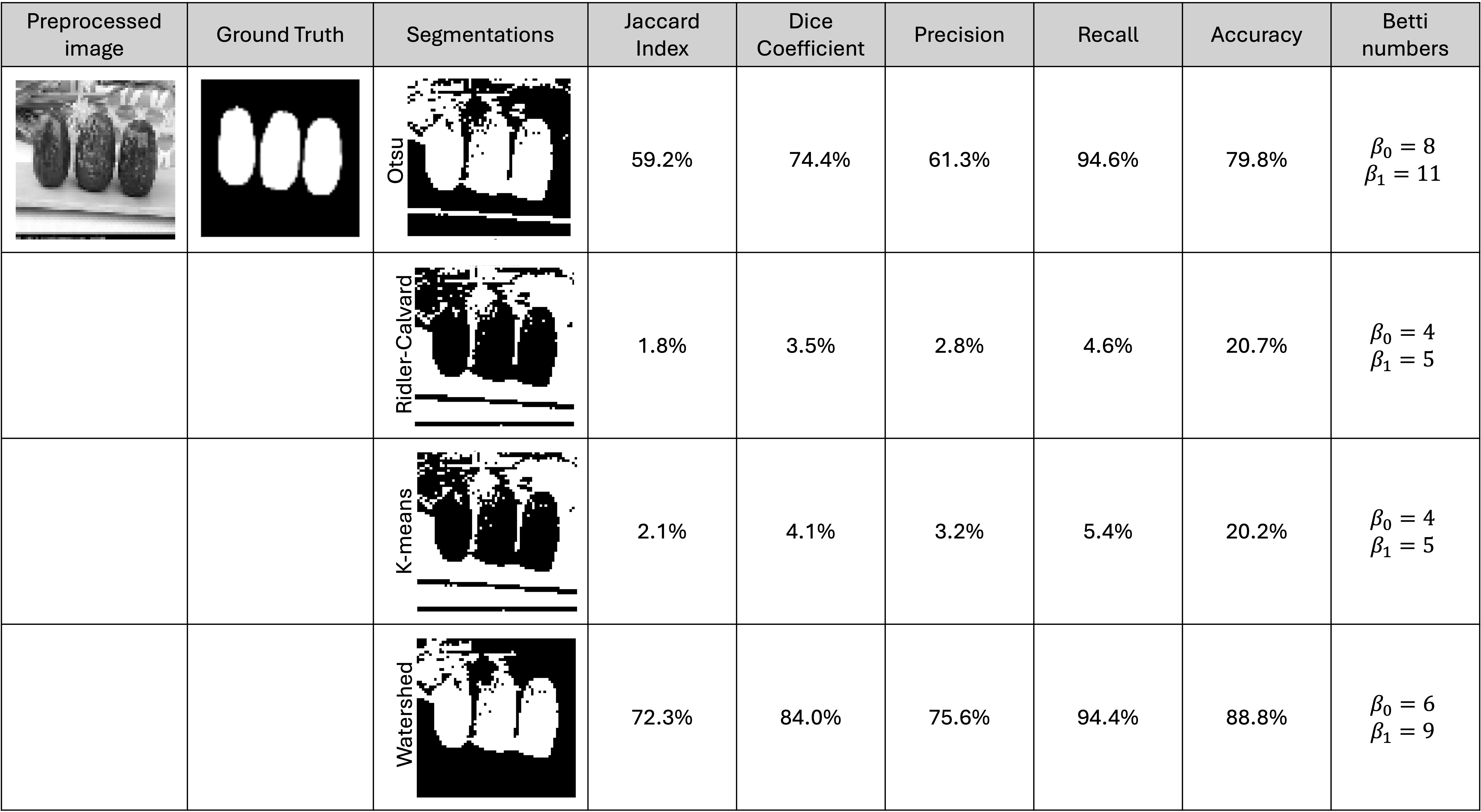}
    \caption{Example 1: Original image and corresponding segmentation masks obtained using different algorithms, with evaluation metrics (Jaccard Index, Dice Coefficient, Precision, Recall, Accuracy) and Betti numbers.}
    \label{fig:result2}
\end{figure}

\begin{figure}[tb]
    \centering
    \includegraphics[width=0.85\linewidth]{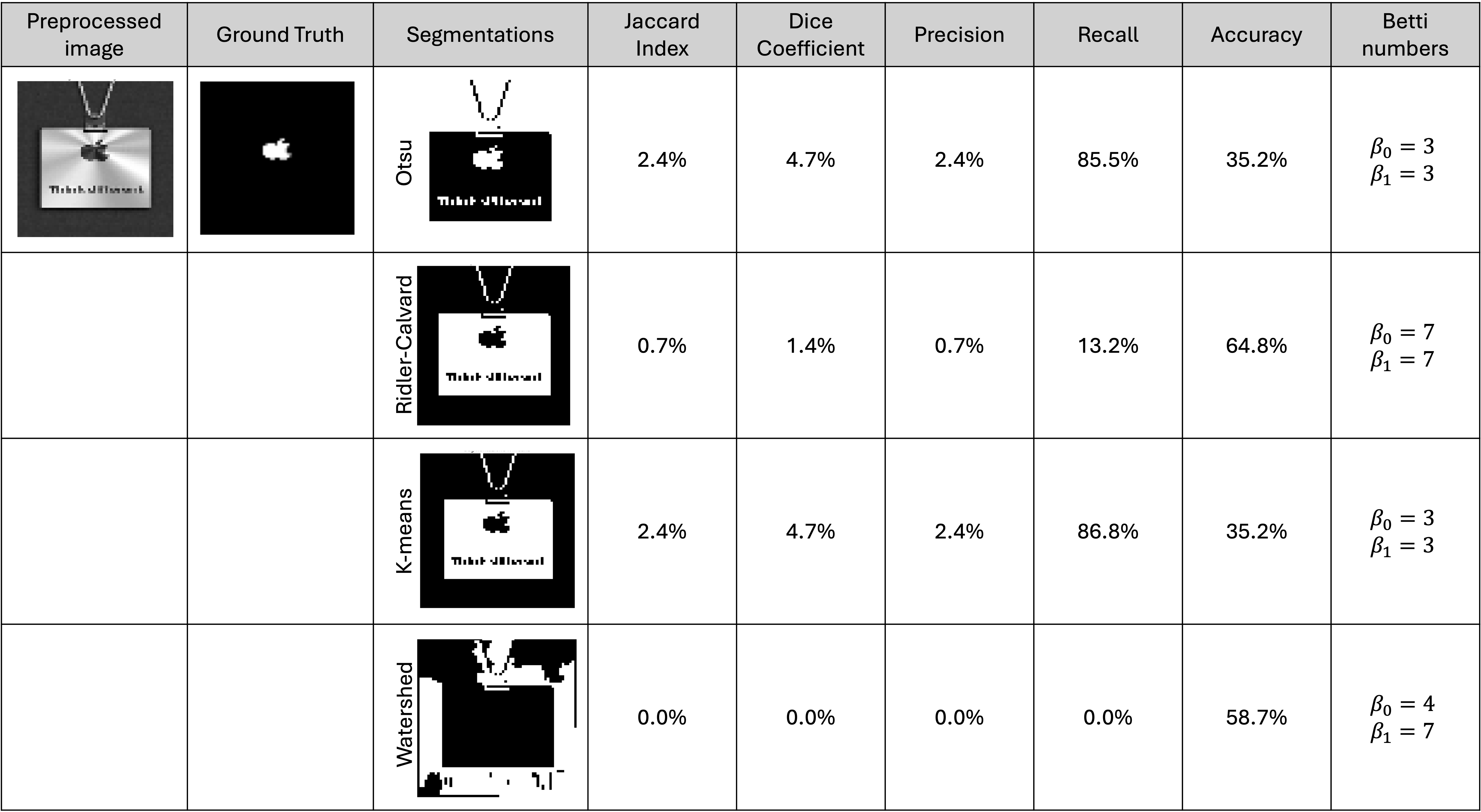}
    \caption{Example 2: Original image and corresponding segmentation masks obtained using different algorithms, with evaluation metrics (Jaccard Index, Dice Coefficient, Precision, Recall, Accuracy) and Betti numbers.}
    \label{fig:result3}
\end{figure}

\begin{figure}[tb]
    \centering
    \includegraphics[width=0.85\linewidth]{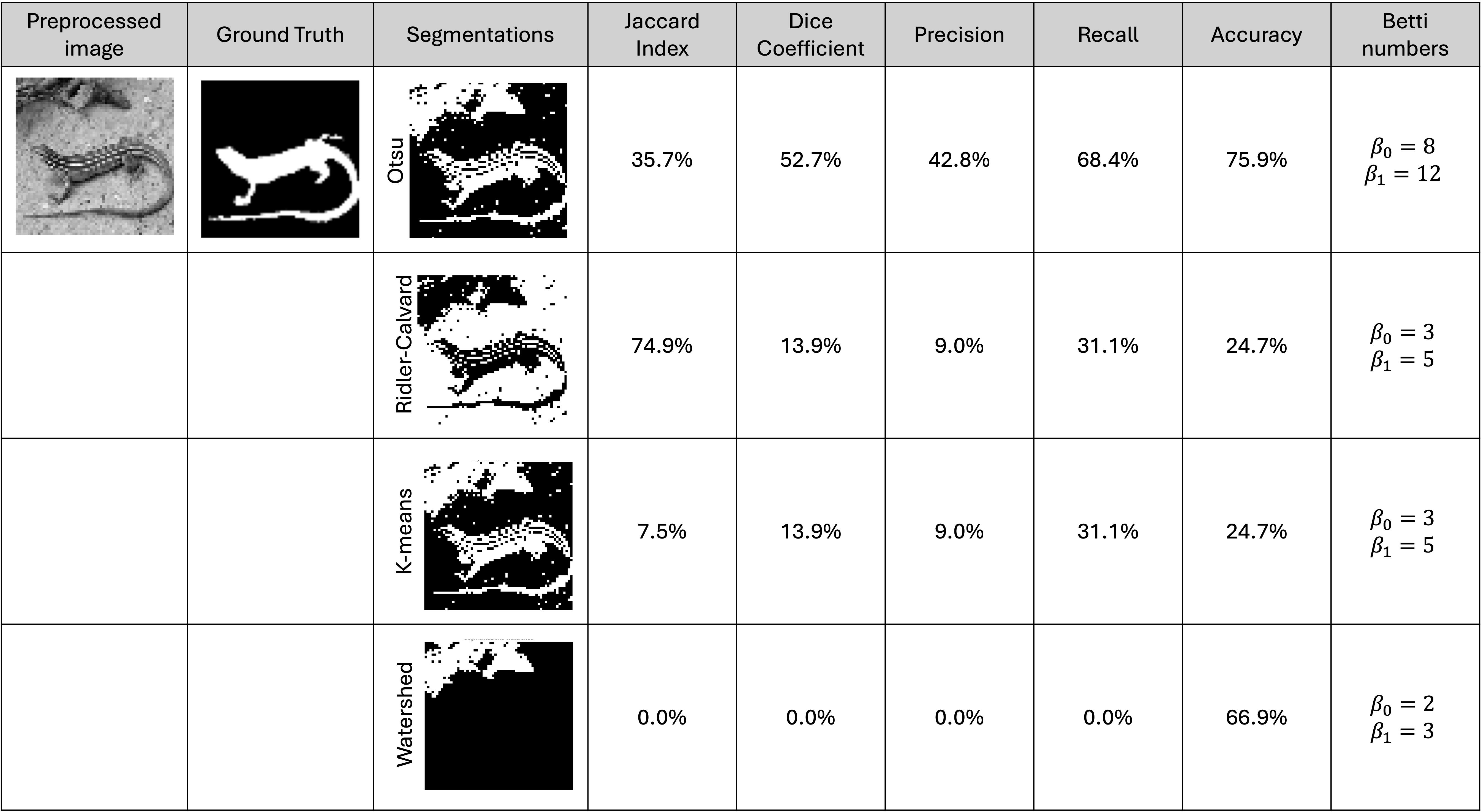}
    \caption{Example 3: Original image and corresponding segmentation masks obtained using different algorithms, with evaluation metrics (Jaccard Index, Dice Coefficient, Precision, Recall, Accuracy) and Betti numbers.}
    \label{fig:result4}
\end{figure}

\begin{figure}[tb]
    \centering
    \includegraphics[width=0.85\linewidth]{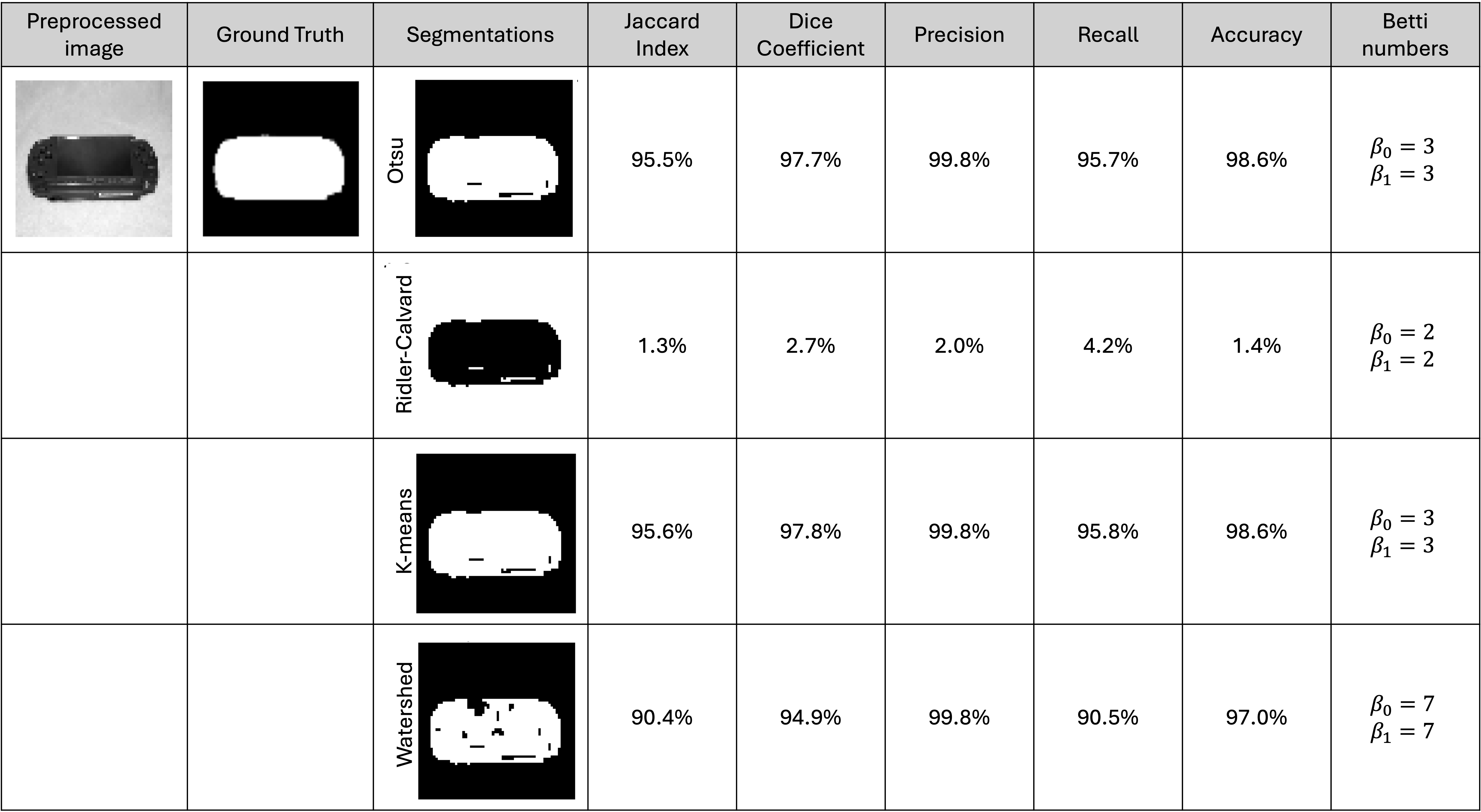}
    \caption{Example 4: Original image and corresponding segmentation masks obtained using different algorithms, with evaluation metrics (Jaccard Index, Dice Coefficient, Precision, Recall, Accuracy) and Betti numbers.}
    \label{fig:result5}
\end{figure}

\begin{figure}[htb]
    \centering
    \includegraphics[width=0.85\linewidth]{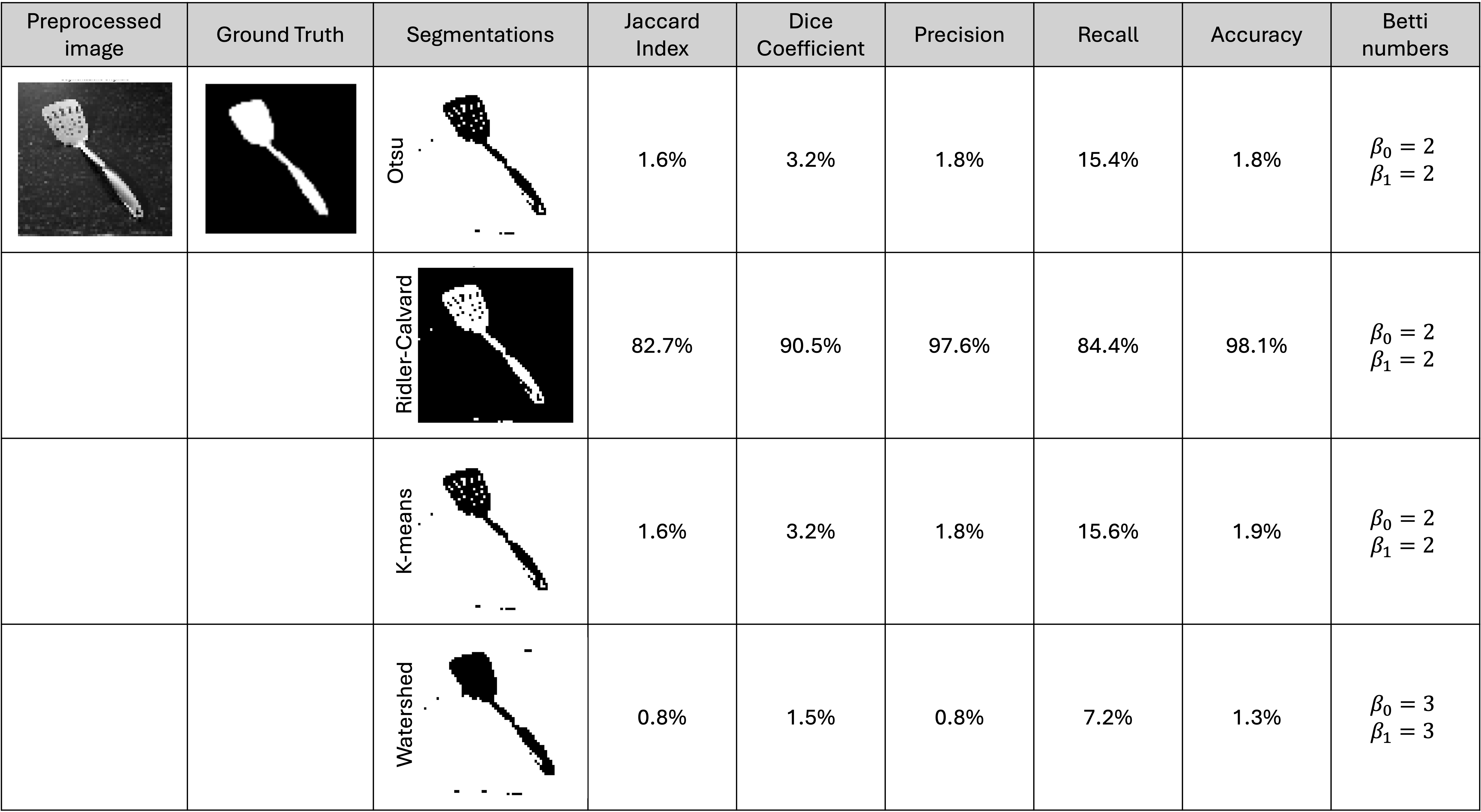}
    \caption{Example 5: Original image and corresponding segmentation masks obtained using different algorithms, with evaluation metrics (Jaccard Index, Dice Coefficient, Precision, Recall, Accuracy) and Betti numbers.}
    \label{fig:result6}
\end{figure}

We want to compare our Jordan-segmentable mask definition with the most commonly employed segmentation metrics in the literature for binary segmentations. Specifically, if we let $P$ denote the set of pixels predicted as foreground (i.e., the predicted segmentation mask) and $G$ denote the set of ground truth foreground pixels (i.e., the ground truth mask), then we will use the following evaluation metrics:
\begin{itemize}
    \item \textbf{Intersection over Union (IoU)} or \textbf{Jaccard Index} measures the overlap between the predicted mask and the ground truth, and is defined as
\[
\mathrm{IoU}(P, G) = \frac{|P \cap G|}{|P \cup G|};
\]

\item \textbf{Precision} measures the proportion of predicted positives that are actually correct:
\[
\mathrm{Precision}(P, G) = \frac{|P \cap G|}{|P|};
\]
\item \textbf{Recall (or Sensitivity)} measures the proportion of ground truth positives that are correctly predicted:
\[
\mathrm{Recall}(P, G) = \frac{|P \cap G|}{|G|};
\]
\item \textbf{Dice Coefficient (or F1 Score)} is the harmonic mean of precision and recall
\[
\mathrm{Dice}(P, G) = \frac{2|P \cap G|}{|P| + |G|};
\]

\item \textbf{Accuracy} measures the proportion of correctly classified pixels (both foreground and background):
\[
\mathrm{Accuracy}(P, G) = \frac{|P \cap G| + |I\setminus (P \cup G)|}{|I|}
\]
where $I \setminus (P \cup G)$ represents the set of true negatives (correctly predicted background pixels).
\end{itemize}

As Figures~\ref{fig:result2} and~\ref{fig:result4} illustrate, the segmentation outputs are highly fragmented, highlighting the limitations of traditional pixel-wise evaluation metrics. While these metrics generally produce high scores, the predicted segmentation masks are not Jordan-segmentable when mask labels (0 and 1) are inverted, resulting in lower values. This is due to a lack of spatial continuity in the segmented regions. Therefore, despite favorable pixel-wise metric values, the resulting segmentations are not acceptable according to the topological assessment criterion derived from the digital Jordan theorem, which requires the presence of a 4-curve to clearly separate the object from the background.

Figure~\ref{fig:result3} showns how the approach based on the definition of a Jordan-segmentable mask, can complement traditional evaluation metrics. As an unsupervised and topology-based criterion,  it effectively assesses segmentations that deviate from the ground truth while preserving topological correctness. This provides an additional, meaningful layer of evaluation beyond conventional approaches.

Finally, Figures~\ref{fig:result5} and~\ref{fig:result6} show that the results obtained through the proposed approach are consistent with those of traditional evaluation metrics provided that the segmentation masks are coherent with the ground truth. However, when the masks are not perfectly aligned with the ground truth (e.g., pixel labels 0 and 1 are inverted), the proposed method can still capture the underlying topological structure. This indicates its ability to provide meaningful and complementary insights, even when conventional metrics cannot.

\section{Conclusion}\label{sec:conclusion}

In this work, we introduced the concept of a Jordan-segmentable mask: a topology-aware notion that is designed to be used as an unsupervised criterion for evaluating binary segmentations. Building on the digital Jordan Curve Theorem and homological invariants, this concept offers a complementary perspective to traditional metrics. While classical measures usually focus on pixel-wise agreement, they can overlook structural issues such as small holes, disconnected components or accidental fragmentation. In contrast, our criterion offers a more principled way to verify whether the foreground and background are separated by a valid $4$-curve.

Experiments conducted on the FSS-1000 dataset, using classical segmentation methods such as Otsu, Ridler–Calvard, K-means, and watershed, illustrate the usefulness of this approach. In most cases, the ground truth masks satisfy the expected topological properties ($\beta_0(S)=1$, $\beta_1(S)=1$, and $\beta_0(I \setminus S)=2$, including affine configurations), demonstrating their natural conformity to the definition of Jordan-segmentable masks. Notably, the topological criterion serves not only to confirm good segmentations, but also to reinterpret results that conventional metrics might deem unsatisfactory by revealing underlying structural correctness that would otherwise remain unseen.

The aim of this study is to highlight the expressive power of the Jordan Curve Theorem in digital image analysis and encourage its use as a structural prior. Looking ahead, several promising avenues emerge, such as experimenting with additional segmentation techniques, extending the concept to multi-class settings and integrating the Jordan-segmentable mask definition directly into segmentation pipelines. This would result in outputs that are not only accurate, but also structurally coherent.

\section*{Acknowledgements}
The authors S.G.D. and N.D.B. are members of the Gruppo Nazionale Calcolo Scientifico - Istituto Nazionale di Alta Matematica (GNCS-INdAM).  S.G.D. is funded by a PhD fellowship within the framework of the Italian “D.M. n. 117, March 2, 2023” - under the National Recovery and Resilience Plan, Msn. 4, Comp. 2, Investment 3.3 - PhD Project “Topological Data Analysis and optimization for industrial processes”, co-supported by “Pirelli Tyre S.p.A.” (CUP H91I23000170007).
This work was supported by INdAM - GNCS Project \textit{MODA: Integrating MOdel-based and DAta-Driven Methods for Multiscale Biological Systems} (CUP E53C24001950001, to S.G.D.).

A.A. was partially supported by PRIN 2022MWPMAB - “Interactions between Geometric Structures and Function Theories” and by GNSAGA of INdAM

\section*{Author Contribution}
Serena Grazia De Benedictis: Conceptualization, Methodology, Software, Validation, Formal analysis, Writing – Original Draft, Writing – Review \& Editing, Visualization.  
Nicoletta Del Buono and Amedeo Altavilla: Conceptualization, Writing – Review \& Editing, Supervision.

\bibliographystyle{plain}  % oppure sn-mathphys-num se disponibile
\bibliography{sn-bibliography} % senza .bib alla fine

@book{graph_theory,
  title     = {Graph Theory with Applications},
  author    = {Bondy, J. A. and Murty, U. S. R.},
  publisher = {Elsevier Science Ltd/North-Holland},
  year      = {1976},
  address   = {London},
  isbn      = {9780444194510}
}

@book{Bredon1993,
  title     = {Topology and Geometry},
  author    = {Bredon, Glen E.},
  publisher = {Springer},
  address   = {New York},
  year      = {1993},
  isbn      = {9781475768480},
  doi       = {10.1007/978-1-4757-6848-0}
}

@book{MoharThomassen,
  title     = {Graphs on Surfaces},
  author    = {Mohar, Bojan and Thomassen, Carsten},
  publisher = {Johns Hopkins University Press},
  address   = {Baltimore, MD},
  year      = {2001},
  isbn      = {9780801866890},
  doi       = {10.56021/9780801866890}
}

@book{simplicial_complex_ref,
  title     = {Algebraic Topology},
  author    = {Maunder, C. R. F.},
  publisher = {Dover Publications},
  address   = {Mineola, NY},
  year      = {1996}
}

@book{binary_segmentation,
  title     = {Image Segmentation: Principles, Techniques, and Applications},
  author    = {Lei, Tao and Nandi, Asoke K.},
  publisher = {Wiley},
  address   = {Hoboken, NJ},
  year      = {2022},
  isbn      = {9781119859048},
  doi       = {10.1002/9781119859048}
}

@book{introduz2,
  title     = {Digital Image Processing},
  author    = {Gonzalez, Rafael C. and Woods, Richard E.},
  edition   = {4},
  publisher = {Pearson},
  address   = {Boston, MA},
  year      = {2018},
  isbn      = {9780133356724}
}

@book{homology_theory_book,
  title     = {Homology Theory},
  author    = {Vick, James W.},
  publisher = {Springer},
  address   = {New York},
  year      = {1994},
  isbn      = {9781461208815},
  doi       = {10.1007/978-1-4612-0881-5}
}

@book{libro_algebraic_topology,
  title     = {An Introduction to Algebraic Topology},
  author    = {Rotman, Joseph J.},
  publisher = {Springer},
  address   = {New York},
  year      = {1998}
}

@article{introduz1,
  title     = {Image segmentation using deep learning: A survey},
  author    = {Minaee, Shervin and Boykov, Yuri and Porikli, Fatih and Plaza, Antonio and Kehtarnavaz, Nasser and Terzopoulos, Demetri},
  journal   = {IEEE Transactions on Pattern Analysis and Machine Intelligence},
  volume    = {44},
  number    = {7},
  pages     = {3523--3542},
  year      = {2021},
  doi       = {10.1109/TPAMI.2021.3059968}
}

@article{introduz3,
  title     = {Machine learning: Algorithms, real-world applications and research directions},
  author    = {Sarker, Imran H.},
  journal   = {SN Computer Science},
  volume    = {2},
  number    = {3},
  pages     = {160},
  year      = {2021},
  doi       = {10.1007/s42979-021-00592-x}
}

@article{introduz4,
  title     = {Object-based convolutional neural network for high-resolution imagery classification},
  author    = {Zhao, Wenji and Du, Shihong and Emery, William J.},
  journal   = {IEEE Journal of Selected Topics in Applied Earth Observations and Remote Sensing},
  volume    = {11},
  number    = {10},
  pages     = {3506--3518},
  year      = {2018},
  doi       = {10.1109/JSTARS.2018.2841929}
}

@article{introduz5,
  title     = {A survey on object detection in optical remote sensing images},
  author    = {Cheng, Gong and Han, Junwei and Lu, Xiang},
  journal   = {ISPRS Journal of Photogrammetry and Remote Sensing},
  volume    = {144},
  pages     = {90--114},
  year      = {2018},
  doi       = {10.1016/j.isprsjprs.2018.06.006}
}

@article{introduz6,
  title     = {A review of traditional and deep learning-based segmentation methods for brain MR images},
  author    = {Yousefi, Mehrdad and Kehtarnavaz, Nasser},
  journal   = {Journal of Digital Imaging},
  volume    = {35},
  pages     = {1123--1136},
  year      = {2022},
  doi       = {10.1007/s10278-022-00648-y}
}

@article{introduz7,
  title     = {Comparison of thresholding techniques for segmentation of leaf diseases},
  author    = {Hasan, Md Morshed and Islam, Md Tariqul},
  journal   = {Egyptian Informatics Journal},
  volume    = {22},
  number    = {2},
  pages     = {115--123},
  year      = {2021},
  doi       = {10.1016/j.eij.2020.02.001}
}

@article{spazio_topologico_4_adj,
  title     = {A Special Topology for the Integers (Problem 5712)},
  author    = {Marcus, D. and Wyse, F.},
  journal   = {The American Mathematical Monthly},
  volume    = {77},
  pages     = {1119},
  year      = {1970},
  doi       = {10.2307/2316121}
}

@article{spazio_topologico_8_adj,
  title     = {Connectivity and consecutivity in digital pictures},
  author    = {Chassery, Jean-Marc},
  journal   = {Computer Graphics and Image Processing},
  volume    = {9},
  pages     = {294--300},
  year      = {1979},
  doi       = {10.1016/0146-664X(79)90043-1}
}

@article{spazio_topologico_6_adj,
  title     = {Comparability Graphs and Digital Topology},
  author    = {Bretto, Alain},
  journal   = {Computer Vision and Image Understanding},
  volume    = {82},
  number    = {1},
  pages     = {33--41},
  year      = {2001},
  doi       = {10.1006/cviu.2000.0901}
}

@article{Rosenfeld_Digital_topology,
  title     = {Digital Topology},
  author    = {Rosenfeld, Azriel},
  journal   = {The American Mathematical Monthly},
  volume    = {86},
  number    = {8},
  pages     = {621--630},
  year      = {1979},
  doi       = {10.1080/00029890.1979.11994873}
}

@article{Rosenfeld_Arcs_and_Curves,
  title     = {Arcs and Curves in Digital Pictures},
  author    = {Rosenfeld, Azriel},
  journal   = {Journal of the ACM},
  volume    = {20},
  number    = {1},
  pages     = {81--87},
  year      = {1973},
  doi       = {10.1145/321738.321745}
}

@article{ROSENFELD_converse_jordan,
  title     = {A converse to the Jordan curve theorem for digital curves},
  author    = {Rosenfeld, Azriel},
  journal   = {Information and Control},
  volume    = {29},
  number    = {3},
  pages     = {292--293},
  year      = {1975},
  doi       = {10.1016/S0019-9958(75)90459-3}
}

@article{Fary_converse_jordan_on_R^2,
  title     = {On a Converse of the Jordan Curve Theorem},
  author    = {Fary, Istvan and Isenberg, Eric M.},
  journal   = {The American Mathematical Monthly},
  volume    = {81},
  number    = {6},
  pages     = {636},
  year      = {1974},
  doi       = {10.2307/2319221}
}

@article{Otsu1979,
  title     = {A Threshold Selection Method from Gray-Level Histograms},
  author    = {Otsu, Nobuyuki},
  journal   = {IEEE Transactions on Systems, Man, and Cybernetics},
  volume    = {9},
  number    = {1},
  pages     = {62--66},
  year      = {1979},
  doi       = {10.1109/TSMC.1979.4310076}
}

@article{RidlerCalvard1978,
  title     = {Picture Thresholding Using an Iterative Selection Method},
  author    = {Ridler, T. W. and Calvard, S.},
  journal   = {IEEE Transactions on Systems, Man, and Cybernetics},
  volume    = {8},
  number    = {8},
  pages     = {630--632},
  DOI = {10.1109/tsmc.1978.4310039},
  year      = {1978}
}

@article{metriche,
  title     = {Towards a guideline for evaluation metrics in medical image segmentation},
  author    = {Müller, Dominik and Soto-Rey, Iñaki and Kramer, Frank},
  journal   = {BMC Research Notes},
  volume    = {15},
  number    = {1},
  pages     = {1--10},
  year      = {2022},
  doi       = {10.1186/s13104-022-06096-y}
}

@article{metriche2,
  title     = {Image segmentation evaluation: a survey of methods},
  author    = {Wang, Zhaobin and Wang, E. and Zhu, Ying},
  journal   = {Artificial Intelligence Review},
  volume    = {53},
  number    = {8},
  pages     = {5637--5674},
  year      = {2020},
  doi       = {10.1007/s10462-020-09830-9}
}

@inproceedings{MacQueen1967,
  title        = {Some Methods for Classification and Analysis of Multivariate Observations},
  author       = {MacQueen, J.},
  booktitle    = {Proceedings of the Fifth Berkeley Symposium on Mathematical Statistics and Probability},
  year         = {1967},
  pages        = {281--297},
  url={https://api.semanticscholar.org/CorpusID:6278891}
}

@inproceedings{BeucherLantuejoul1979,
  title        = {Use of Watersheds in Contour Detection},
  author       = {Beucher, S. and Lantu{\'e}joul, C.},
  booktitle    = {Proceedings of the International Workshop on Image Processing},
  year         = {1979},
  pages        = {17--21},
  url={https://api.semanticscholar.org/CorpusID:59652065}
}

@inproceedings{dataset_FSS-1000,
  title        = {FSS-1000: A 1000-Class Dataset for Few-Shot Segmentation},
  author       = {Li, Xiang and Wei, Tianhan and Chen, Yau Pun and Tai, Yu-Wing and Tang, Chi-Keung},
  booktitle    = {Proceedings of the IEEE/CVF Conference on Computer Vision and Pattern Recognition (CVPR)},
  year         = {2020},
  pages        = {294--303},
  doi          = {10.1109/CVPR42600.2020.00294},
  address      = {Seattle, WA}
}

@inbook{distanze_PB,
  title      = {Computational Topology: An Introduction},
  author     = {Rote, G{\"u}nter and Vegter, Gert},
  booktitle  = {Effective Computational Geometry for Curves and Surfaces},
  publisher  = {Springer},
  address    = {Berlin},
  pages      = {277--312},
  year       = {2006},
  doi        = {10.1007/978-3-540-33259-6_7}
}

@misc{lavoro,
  title      = {Rethinking Semantic Segmentation Evaluation for Explainability and Model Selection},
  author     = {Zhang, Yuxiang and Mehta, Sachin and Caspi, Anat},
  year       = {2021},
  note       = {arXiv:2101.08418},
  doi        = {10.48550/arXiv.2101.08418}
}

\end{document}